\def\year{2021}\relax
\documentclass[letterpaper]{article} 
\usepackage{aaai21}  
\usepackage{times}  
\usepackage{helvet} 
\usepackage{courier}  
\usepackage[hyphens]{url}  
\usepackage{graphicx} 
\usepackage{natbib}  
\usepackage{caption} 
\usepackage{url}
\usepackage{times}
\usepackage{epsfig}
\usepackage{graphicx}
\usepackage{amsmath}
\usepackage{bm}
\usepackage{amssymb}
\usepackage{color}
\usepackage{mathtools,xparse}
\usepackage{enumitem}
\usepackage{multirow}
\usepackage{rotating}
\usepackage[ruled]{algorithm2e}
\usepackage[utf8]{inputenc} 
\usepackage{url}            
\usepackage{booktabs}       
\usepackage{amsfonts}       
\usepackage{nicefrac}       
\usepackage{microtype}      
\usepackage{tabularx}
\usepackage{subcaption}
\usepackage{arydshln}
\usepackage{siunitx}
\usepackage{soul}
\title{Classification by Attention: \\ Scene Graph Classification with Prior Knowledge}

\author{%
	Sahand Sharifzadeh,$^{1}$ 
	Sina Moayed Baharlou,$^{2}$
	Volker Tresp$^{1,3}$ \\
}
\affiliations{
	$^{1}$Ludwig Maximilian University of Munich, 
	$^{2}$Sapienza University of Rome,
	$^{3}$Siemens AG\vspace{.7em} \\   
	\texttt{sharifzadeh@dbs.ifi.lmu.de}
}

\makeatletter
\newcommand*{\rom}[1]{\expandafter\@slowromancap\romannumeral #1@}
\makeatother
\begin{document}

	\maketitle
	\begin{abstract}
		A major challenge in scene graph classification is that the appearance of objects and relations can be significantly different from one image to another. Previous works have addressed this by relational reasoning over all objects in an image or incorporating prior knowledge into classification. 
		Unlike previous works, we do not consider separate models for perception and prior knowledge. Instead, we take a multi-task learning approach, where we implement the classification as an attention layer. This allows for the prior knowledge to emerge and propagate within the perception model. By enforcing the model also to represent the prior, we achieve a strong inductive bias. 
		We show that our model can accurately generate commonsense knowledge and that the iterative injection of this knowledge to scene representations leads to significantly higher classification performance. Additionally, our model can be fine-tuned on external knowledge given as triples. When combined with self-supervised learning and with 1\% of annotated images only, this gives more than 3\% improvement in object classification, 26\% in scene graph classification, and 36\% in predicate prediction accuracy.
	\end{abstract}
	
	\section{Introduction}
\begin{figure}[t]
	\begin{center}
		\includegraphics[width=0.42\textwidth]{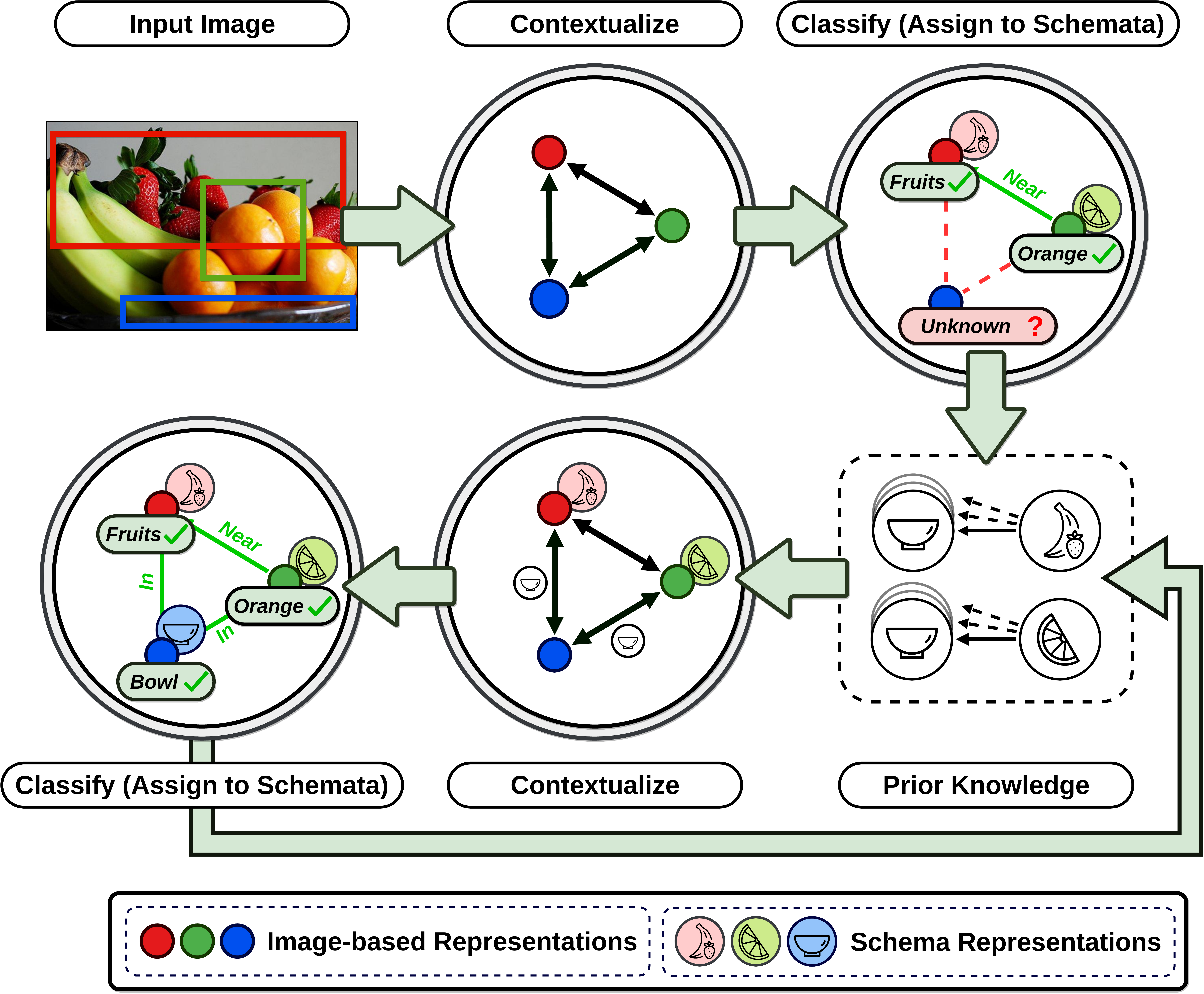}
	\end{center}
	\caption{An example of scene graph classification where the \texttt{Bowl} lacks sufficient visual input. Top right is the initially predicted graph from visual inputs only. Bottom left is the prediction of our model after considering both image-based representations and schematic prior knowledge about \texttt{Fruits} and \texttt{Oranges}. The long arrow near the bottom indicates recursion.
	}	\label{fig_intro}
\end{figure}

	Classifying objects and their relations in images, also known as scene graph classification, is a fundamental task in scene understanding and can play an essential role in applications such as recommender systems, visual question answering and decision making. Scene graph (SG) classification methods typically have a perception model that takes an image as input and generates a graph that describes the given image as a collection of (\texttt{head, predicate, tail}). One of the main challenges that current models face is diverse appearances of objects and relations across different images. This can be due to variations in lighting conditions, viewpoints, object poses, occlusions, etc.
	For example, the \texttt{Bowl} in Figure \ref{fig_intro} is highly occluded and has very few image-based features. Therefore, a typical perception model fails to classify it. One approach to tackle this problem is to collect supportive evidence from neighbors before classifying an entity. This can be done, for example, by message passing between all the image-based object representations in an image, using graph convolutional neural networks (GCN)~\citep{kipf2016semi} or LSTMs~\citep{hochreiter1997long}. The main issue with this approach is the combinatorial explosion of all possible \textit{image-based} neighbor representations\footnote{For a more detailed probabilistic analysis of this issue, refer to the section \textit{GCN vs. Prior Model: A matter of inductive biases.}}.

	A current theory in cognitive psychology states that humans solve this challenge by reasoning over the pre-existing representations of neighboring objects instead of relying on the perceptual inputs only~\citep{piaget1923langage}; philosophers often argue that humans have a form of mental representation for objects and concepts~\citep{kant2011kritik}. These representations do not depend on a given image but are rather \textit{symbol-based}. There are different opinions on how these representations come to be.
	Piaget called these representations \textit{schema} (plural \textit{schemata}) and suggested that we acquire them in our earlier perceptions.
	When an object is being perceived, the mind assigns it to a schema in a process called \textit{assimilation}. By relational reasoning over schemata, assimilation helps to predict the facts surrounding the observation~\citep{arbib1992schema}\footnote{We avoid discussing the differences between Kant's and Piaget's schema for now.}. 
	
	Nevertheless, learning and utilizing prior knowledge is still a significant challenge in AI research. Recently, ~\citet{zellers2018neural} and~\citet{chen2019knowledge,chen2019learning} proposed to correct the SG prediction errors by comparing them to the co-occurrence statistics of triples in the training dataset or an external source. The statistics can, for example, suggest that it is common to see \texttt{(Fruits, in, Bowls)}. Furthermore, instead of relying on simple co-occurrence statistics, one can create a prior model with knowledge graph embeddings (KGE)~\citep{nickel2016review} that can generalize beyond the given triples~\citep{baier2017improving,baier2018improving, hou2019relational}.
	KGEs typically consist of an \textit{embedding matrix}, that assigns a vector representation to each entity, and an \textit{interaction function} that predicts the probability of a triple given the embeddings of its head, predicate, and tail. 
	This allows KGE models to generalize to unseen relations. For example, \texttt{Man} and \texttt{Woman} will be given similar embeddings since they appear in similar relations during the training. As a result, if the model observes \texttt{(Woman, rides, Horse)}, it can generalize, for example, to \texttt{(Man, rides, Camel)}. 
	
	However, in the described approaches, unlike Piaget's schemata, the perception and the prior are treated naively as independent components; they are trained separately and from different inputs (either from images or from triples), and their predictions are typically fused in the probability space, e.g., by multiplication. Other than requiring redundant parameters and computations, this makes the prior model agnostic to the image-based knowledge and the perception model agnostic to the prior knowledge. For example, the collection of triples might contain \texttt{(Woman, rides, Horse)} but have no triples regarding a \texttt{Donkey}. While the images can represent the visual similarities of a \texttt{Horse} to a \texttt{Donkey}, the triples lack this information. If we train a prior model purely based on the triples, the model fails to generalize. We can avoid this by training the prior model from a combination of triples and images. As for another example, in Figure \ref{fig_intro}, the prior model might suggest \texttt{(Fruits, in, Bowl)} but it might also suggest \texttt{(Fruits, areOn, Trees)}. To decide between the two, one should still consider the visual cues from the given image.

\begin{figure}
		\centering
		\includegraphics[width=0.38\textwidth]{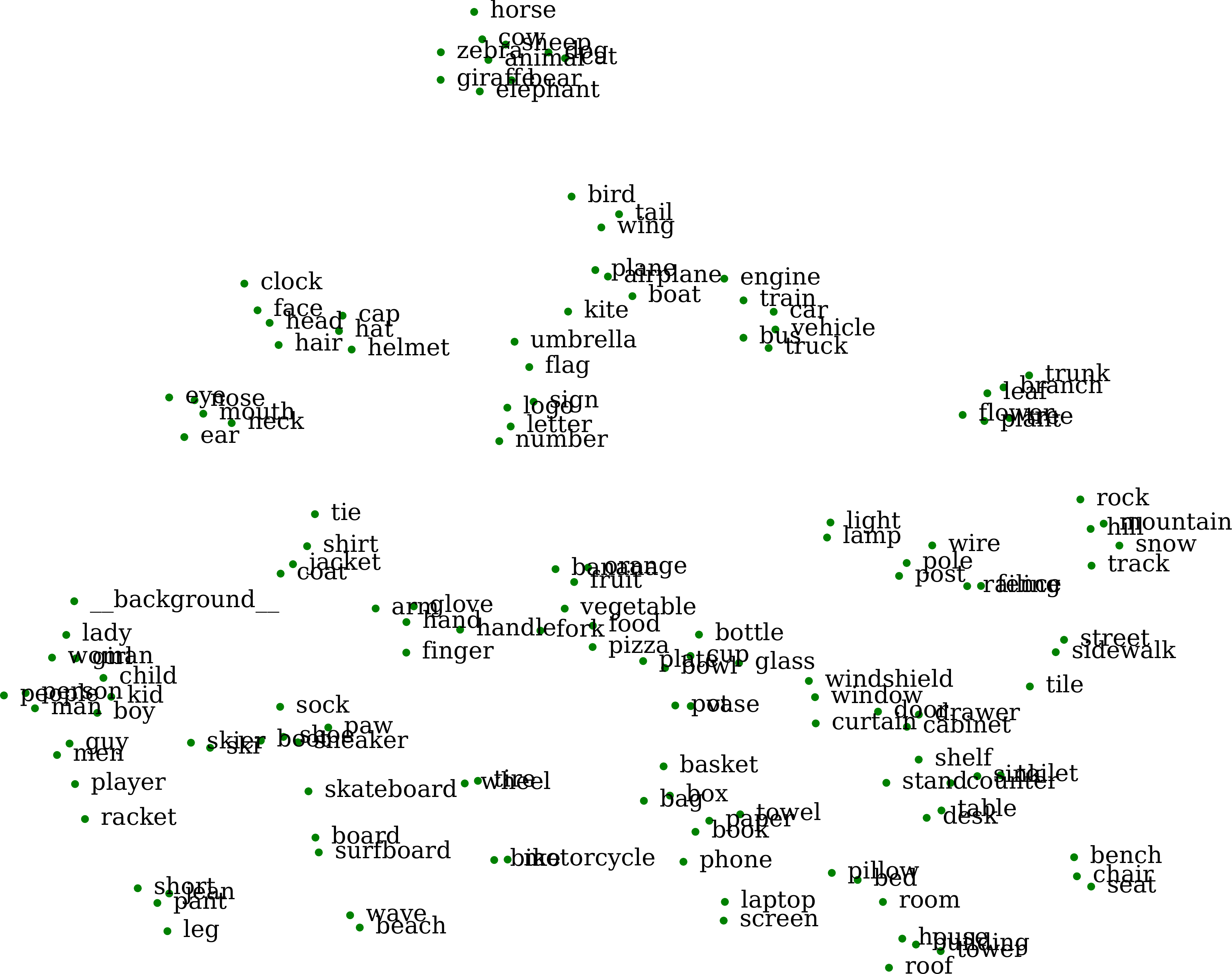}
		\caption{t-SNE visualization of the object classification weights that have been trained on top of contextualized image-based representations. Entities that appear similar to each other, or participate in similar relations, have a closer semantic affinity. This enables link prediction similar to Knowledge Graph Embeddings, and leads to generalization.}\label{fig_tsne_schema}
	\end{figure}

	To address these shortcomings, we entangle the perception and prior in a single model with shared parameters trained by multi-task learning. Therefore, instead of training a separate embedding matrix for a prior model, we exploit the perception model's classification layer; when we train a classification layer on top of contextualized image-based representations, the classification weights capture both relational and image-based class embeddings (Refer to Figure~\ref{fig_tsne_schema}).
	Unfortunately, the classification's common realization as a fully connected layer does not allow us to feed these \textit {network weights} to an interaction function.
	To this end, we employ a more general formulation of classification as an attention layer instead. In this layer, the extracted image-based and contextually enriched representations attend to trainable schema embeddings of all classes such that (a) the attention coefficients are the classification scores (we enforce this by applying a classification loss on the attention outputs), and (b) the attention values carry the prior knowledge that is injected into the image-based representations.
	
	Furthermore, instead of training a separate interaction function for the prior model, we exploit the message passing function that we already have available in the perception model; after fusing the schemata and the image-based object-representations, we contextualize and classify the representations again. Other than computational efficiency, this has the advantage that the image-based object representations and the schemata are combined in the embedding space rather than the probability space. 
	
	We train schemata using the Visual Genome~\citep{krishna2017visual} dataset. We show that our model can accurately generate the captured commonsense knowledge and that iterative injection of this knowledge leads to significantly higher classification accuracy. Additionally, we draw from the recent advancements in self-supervised learning and show that the schemata can be trained with only a fraction of labeled images. This allows us to fine-tune the perception model without any additional images; instead, we can use a knowledge base of hand-crafted or external triples and train with their \textit{mental images} (schemata).
	As a result, compared to the self-supervised baseline, and with 1\% of the training data, our model achieves more than 3\% improvement in object classification, 26\% in scene graph classification, and 36\% in predicate prediction; an accuracy that is almost equal to when using 100$\%$ of the labeled images.

	\section{Related Works}
	While the concept of schemata can be applied to any form of perceptual processing, and there are recent deep learning models of action schemata~\citep{kansky2017schema,goyal2020object}, we focus on the figurative schemata in the visual scene understanding domain. 
    Even though there has been a body of related research outside the scene graph domain~\citep{wu2014hierarchical,deng2014large,hu2016learning,hu2017labelbank,santoro2017simple}, research in this field was accelerated mainly after the release of Visual Relation Detection (VRD)~\citep{lu2016visual} and the Visual Genome~\citep{krishna2017visual} datasets. ~\citet{baier2017improving,baier2018improving} proposed the first KG-based model of prior knowledge that improves SG classification. VTransE~\citep{zhang2017visual} proposed to capture relations by applying the KGE model of TransE~\citep{bordes2013translating} on the visual embeddings.~\citet{yu2017visual} employed a teacher-student model to distill external language knowledge. Iterative Message Passing~\citep{xu2017scene}, Neural Motifs~\citep{zellers2018neural} (NM), and Graph R-CNN~\citep{yang2018graph} used RNNs and graph convolutions to propagate image context.~\citet{tang2019learning} exploited dynamic tree structures and~\citet{chen2019counterfactual} proposed a method based on multi-agent policy gradients.~\citet{sharifzadeh2019improving} employed the predicted pseudo depth maps of the images in addition to the 2D information.
	In general, scene graph classification methods are closely related to KGE models~\citep{nickel2011three,nickel2016review}. For an extensive discussion on the connection between perception, KG models, and cognition, refer to~\citep{tresp2019model,tresp2020tensor}.
	The link prediction in KGEs arises from the compositionality of the trained embeddings. Other forms of compositionality in neural networks are discussed in other works such as~\citep{montufar2014number}.
	In this work, we introduce assimilation, which strengthens the representations within the neural network's causal structure, addressing an issue of neural networks raised by~\citet{fodor1988connectionism}. In general, some of the shortcomings that we address in this work have been recently argued by ~\citet{bengio2017consciousness,marcus2018deep}.
	\begin{figure*}
		\begin{center}
			\includegraphics[width=0.77\textwidth]{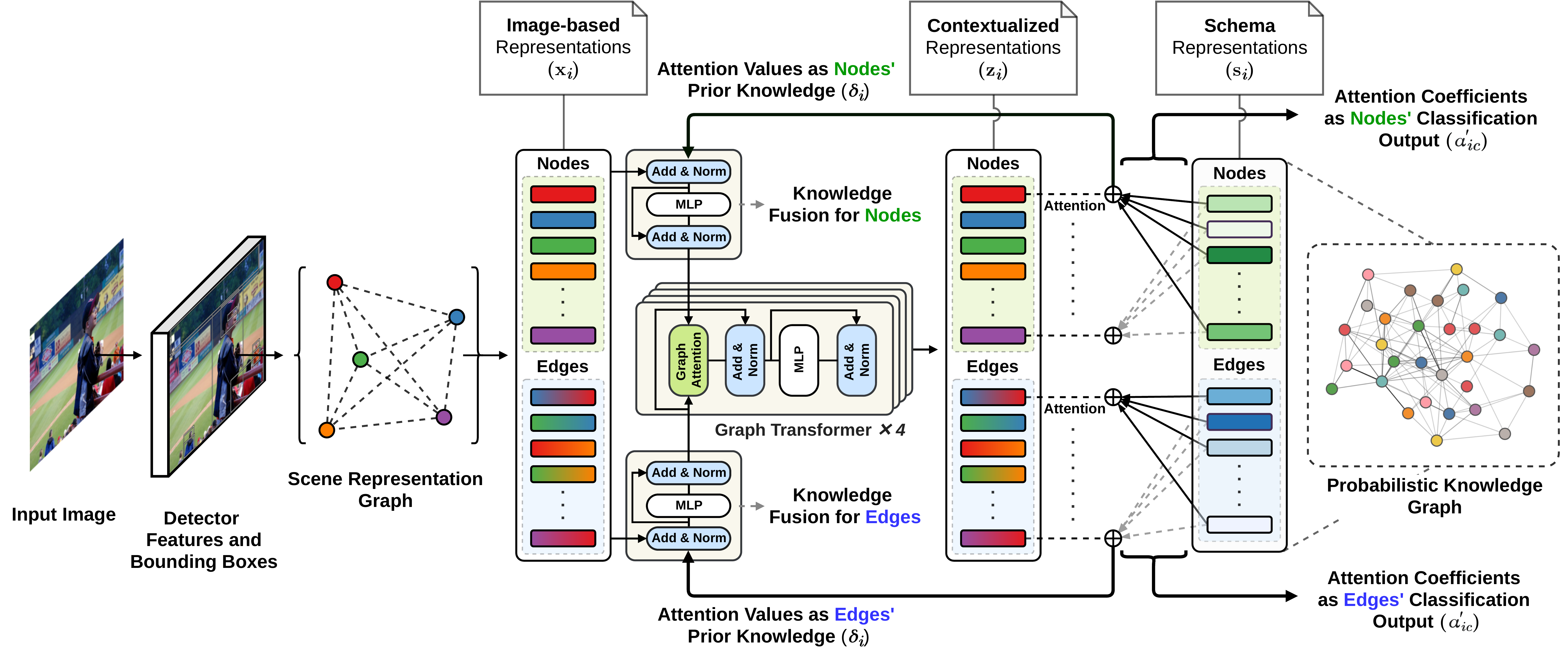}
		\end{center}
		\caption{We formulate the classification as attention layer between object and schema representations. Contextualizing image-based object representations before classification encourages the schemata to learn \textit{image-based relational} prior knowledge. As a result, the attention values that are injected from the schemata to scene representations and then propagated. In this way, they enrich the image-based representations with prior knowledge. Additionally, the interactions between schemata can reconstruct the probabilistic knowledge graph (right).}\label{arch_img}
	\end{figure*}
	\section{Methods}\label{methods}
	In this section we describe our method. In summary, after an initial classification step, we combine the image-based representations with the schema of their predicted class. We then proceed by contextualizing these representations and collecting supportive evidence from neighbors before re-classifying each entity (Ref. Figure \ref{fig_intro}). 	
	In what follows, bold lower case letters denote vectors, bold upper case letters denote matrices, and the letters denote scalar quantities or random variables. We use subscripts and superscripts to denote variables and calligraphic upper case letters for sets.
	\subsection{Definitions}
	Let us consider a given image $\mathbf{I}$ and a set of $n$ bounding boxes $\mathcal{B} = \{\mathbf{b}_i\}_{i=1}^n$, $\mathbf{b}_i=[b^x_i,b^y_i,b^w_i,b^h_i]$, such that $[b^x_i,b^y_i]$ are the coordinates of $\mathbf{b}_i$ and $[b^w_i,b^h_i]$ are its width and height. We build a \textbf{Scene Representation Graph}, $\textit{SRG} = \{\mathcal{V}, \mathcal{E}\}$ as a structured presentation of the objects and predicates in $\mathbf{I}$. $\mathcal{X}^o = \{\mathbf{x}_i^o\}_{i=1}^n$, $\mathbf{x}^o_i \in \mathbb{R}^d$ denote the features of object nodes and $\mathcal{X}^p= \{\mathbf{x}_i^p\}_{i=1}^m$, $\mathbf{x}^p_i \in \mathbb{R}^d$ denote the features of predicate nodes\footnote{Similar to~\citep{yang2018graph, koncel2019text}, we consider each object node as direct neighbors with its predicate nodes and each predicate node as direct neighbors with its head and tail object nodes.}. Each $\mathbf{x}^o_i$ is initialized by a pooled image-based object representation, extracted by applying VGG16~\cite{simonyan2014very} or ResNet-50~\citep{he2016deep} on the image contents of $\mathbf{b}_i$. Each $\mathbf{x}^p_i$ is initialized by applying a two layered fully connected network on the relational position vector $\mathbf{t}$ between a head $i$ and a tail $j$ where $\mathbf{t} = [t_x, t_y, t_{w}, t_{h}]$, 
	$
	t_x = (b^x_{i} - b^x_{j})/{b^w_i}_{j}, t_y = (b^y_{i} - b^y_{j})/b^h_{j}, t_{w} = \log(b^w_{i}/b^w_{j}), t_{h} = \log(b^h_{i}/b^h_{j})
	$. The implementation details of the networks are provided in the Supplementary.
	
	\textbf{Scene graph classification} is the mapping of each node in scene representation graph to a label where each object node is from the label set $\mathcal{C}^o$ and each predicate node from $\mathcal{C}^p$. The resulting labeled graph is a set of triples referred to as the \textbf{Scene Graph}. We also define a \textbf{Probabilistic Knowledge Graph} ($\textit{PKG}$) as a graph where the weight of a triple is the expected value of observing that relation given the head and tail classes and regardless of any given images\footnote{Note that while typical knowledge graphs such as Freebase are based on object instances, given the nature of our image dataset, we focus on classes.}. Later we will show that our model can accurately generate the PKG, i.e., the commonsense that is captured from perceptions during training.
	
	In what comes next, $\mathbf{x}^o_i$ and $\mathbf{x}^p_i$ are treated identically except for classification with respect to $\mathcal{C}^o$ or $\mathcal{C}^p$. Therefore, for a better readability, we only write $\mathbf{x}_i$.
	\subsection{Contextualized Scene Representation Graph}
	We obtain contextualized object representations $\mathbf{z}_i$ by applying a graph convolutional neural network, on $\textit{SRG}$. We sometimes refer to this module as our \textit{interaction function}. We use a Graph Transformer as a variant of the Graph Network Block~\citep{battaglia2018relational,koncel2019text} with multi-headed attentions as
	\begin{align}
	\mathbf{m}_{i}^{\mathcal{N}(i)}=\frac{1}{K}\sum_{k=1}^{K}\sum_{j\in\mathcal{N}(i)}\alpha_{ij}^{(l,k)}\mathbf{W}^{(l,k)}\mathbf{z}^{(l, t)}_{j}
	\end{align}
	\begin{align}
	{\mathbf{z}'}_{i}^{(l)}=\textit{LN}(\mathbf{z}^{(l, t)}_{i} + \mathbf{m}_i^{\mathcal{N}_{in}(i)} + \mathbf{m}_i^{\mathcal{N}_{out}(i)})
	\end{align}
	\begin{align}
	\mathbf{z}_{i}^{(l+1, t)} = \textit{LN}({\mathbf{z}'}_{i}^{(l)} + f({\mathbf{z}'}_{i}^{(l)}))\label{eq_context},
	\end{align}
	where $\mathbf{z}_i^{(l, t)}$ is the embedding of node $i$ in the $l$-th graph convolution layer and $t$-th assimilation. In the first layer $\mathbf{z}_i^{(0,t)} = \mathbf{x}_i$. \textit{LN} is the layer norm~\citep{ba2016layer}, $K$ is the number of attentional heads and $\mathbf{W}^{(l,k)}$ is the weight matrix of the $k$-th head in layer $l$. $\mathcal{N}(i)$ represent the set of neighbors, which are either incoming ${\mathcal{N}_{in}(i)}$ or outgoing $\mathcal{N}_{out}(i)$. $f(.)$ is a two layered feed-forward neural network with Leaky ReLU non-linearities between each layer.  $\alpha_{ij}^{(l,k)}$ denotes the attention coefficients in each head and is defined as
	\begin{align}
	e_{ij}^{(l,k)} &=\sigma(\mathbf{h}^{(l,k)}\cdot [\mathbf{z}_i^{(l)}||\mathbf{W}^{(l,k)}\mathbf{z}^{(l)}_{j}])\\
	\alpha_{ij}^{(l,k)}&=\frac{\exp(e_{ij}^{(l,k)})}{\sum_{q\in \mathcal{N}(i)}^{}\exp(e_{iq}^{(l,k)})}
	\end{align}
	with $\mathbf{h}^{(l,k)}$ as a learnable weight vector and $||$ denoting concatenation. $\sigma$ is the Leaky ReLU with the slope of $0.2$.
	\subsection{Schemata}\label{subseq_assm}
	We define the schema of a class $c$ as an embedding vector $\mathbf{s}_c$. We realize object and predicate classification by an attention layer between the contextualized representations and the schemata such that the classification outputs  $\alpha'_{ic}$ are computed as the attention coefficients between $\mathbf{z}_i$ and $\mathbf{s}_c$ as
	\begin{equation}
	\alpha'_{ic}= \mathrm{softmax}(a(\mathbf{z}_i^{(L, t)}, \mathbf{s}_c))\label{eq_class}
	\end{equation}	
	where, $a(.)$ is the attention function that we implement as the dot-product between the input vectors, and $\mathbf{z}_i^{(L, t)}$ is the output from the last ($L$-th) layer of the Graph Transformer. The attention values $\bm{\delta}_i$ capture the schemata messages as
	\begin{equation}
	\bm{\delta}_i = \sum_{c\in\mathcal{C}}\alpha'_{ic}
	\mathbf{s}_{c}\label{eq_msg}
	\end{equation}
	and we inject them back to update the scene representations as
	\begin{equation}
	\mathbf{u}_i = \textit{LN}(\mathbf{x}_i + \bm{\delta}_i)\label{eq_fusion}
	\end{equation}
	\begin{equation}
	\mathbf{z}_{i}^{(0, t+1)} = \textit{LN}(\mathbf{u}_i + g(\mathbf{u}_i))\label{eq_asm}
	\end{equation}
	where $g(.)$ is a two-layered feed-forward network with Leaky ReLU non-linearities. Note that we compute $\mathbf{u}_i$ by fusing the attention values with the \textit{original image features} $\mathbf{x}_i$. Therefore, the outputs from previous Graph Transformer layers will not be accumulated, and the original image-based features will not vanish. 
	
	We define \textit{assimilation} as the set of computations from $\mathbf{z}_i^{(L, t)}$ to $\mathbf{z}_i^{(L, t+1)}$. This includes the initial classification step (Eq. \ref{eq_class}), fusion of schemata with image-based vectors (Eq. \ref{eq_asm}) and the application of the interaction function on the updated embeddings (Eq. \ref{eq_context}). We expect to get refined object representations after the assimilation. Therefore, we assimilate several times such that after each update of the classification results, the priors are also updated accordingly. During training, and for each step of assimilation, we employ a supervised attention loss, i.e. categorical cross entropy, between the one-hot encoded ground truth labels and $\alpha'_{ic}$. This indicates a multi-task learning strategy where one task (for the first assimilation) is to optimize for $P(y_q | x_1, ..., x_{\theta})$, with $x_q$ as a random variable representing the image-based features of $q$, $y_q$ as the label, and $\theta = m + n$. The other set of tasks is to optimize for $P(y^{t+1}_q | x^{t}_1, ..., x^{t}_\theta, y^{t}_1, ..., y^{t}_\theta)$. We refer to the first task as \textbf{IC}, for \textbf{I}mage-based \textbf{C}lassification and to the second set of tasks as \textbf{ICP} for \textbf{I}mage-based \textbf{C}lassification with \textbf{P}rior knowledge. Note that even when no images are available, we can still train for the \textit{ICP} from a collection of external or hand-crafted triples by setting $\mathbf{x}_i = \vec{0}$, $\alpha'_{ic} = onehot(c_i)$, and starting from Eq \ref{eq_msg}. When we solve for both tasks, to stabilize the training, such that it does not diverge in the early steps (when the predictions are noisy), we train our model by a modified form of scheduled sampling~\citep{bengio2015scheduled} such that in Eq \ref{eq_msg}, we gradually replace up to $10$ percent of the randomly sampled false negative predictions $\alpha'$ with the their true labels. This prevents the model to over-fit a specific error rate from a previous assimilation, and encourages it to generalize beyond the number of assimilations that it has been trained for. 
	
	\subsection{GCN vs. Prior Model: A matter of inductive biases}\label{subseq_gcnvsassm}
	Typical GCNs, such as the Graph Transformer, take the features derived from each bounding box as input, apply non-linear transformations and propagate them to the neighbors in the following layers. 
	Each GCN layer consists of fully connected neural networks. Therefore, \textit{theoretically} they can also model and propagate prior knowledge that is \textit{not} visible in bounding boxes. However, experimental results of previous works (and also this work) confirm that explicit modeling and propagation of prior knowledge (\textit{ICP}) can still improve the classification accuracy. Why is that the case?

	\begin{figure}
		\begin{center}
			\includegraphics[width=0.47\textwidth]{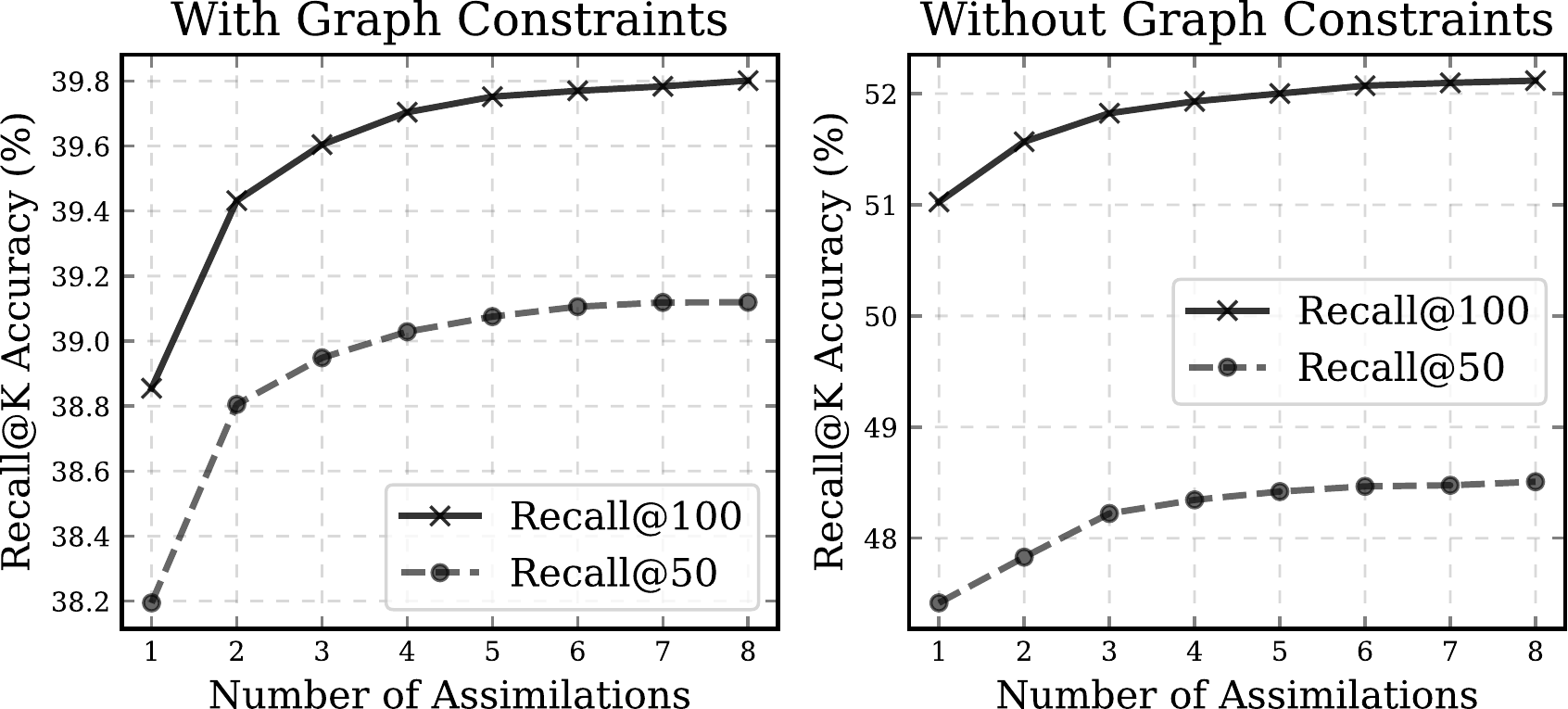}
		\end{center}
		\caption{The results of our ablation studies. We study the effect of each assimilation in scene graph classification. Note that the model has been trained for only 4 assimilations yet it can generalize, thanks to the scheduled sampling strategy.}\label{fig_abl}
	\end{figure}
	Let us consider the following. According to the the universal approximation theorem~\citep{csaji2001approximation}, when we solve for \textit{IC} as $P(y_k | x_1, ..., x_o)$, our model might learn to capture a desired form of $P(y_k | x_1, ..., x_o, y_1, ..., y_o)$. However, in practice, the learning algorithm does not always find the best function. Therefore, we require the proper inductive biases to guide us through the learning process. As ~\citet{caruana1997multitask} puts: \textit{"Multitask Learning is an approach to inductive transfer that improves generalization by using the domain information contained in the training signals of related tasks as an inductive bias. It does this by learning tasks in parallel while using a shared representation; what is learned for each task can help other tasks be learned better"}.
	For example, in the encoder-decoder models for machine translation e.g. Transformers~\citep{vaswani2017attention}, the prediction is often \textit{explicitly} conditioned not just on the encoded inputs but also on the decoded outputs of the previous tokens. Therefore, the decoding in each step can be interpreted as $P(y_k | x_1, ..., x_o, y_1, ..., y_{k-1})$. Note that the previous predictions such as $y_1$, cannot benefit from the future predictions $\{y_2, ...,y_o\}$. However, in our model, we provide an explicit bias towards utilizing predictions in \textit{all indices}. In fact, our model can be interpreted as an encoder-decoder network, where the decoder consists of multiple decoders. Therefore, the decoding depends not just on the encoded image features but also on the previously decoded outputs. In other words, by injecting schema embeddings, as embeddings that are trained over \textit{all images}, we impose the bias to propagate \textit{what is not visible in the bounding box}.
	As will be shown later, we can train for \textit{ICP} and \textit{IC} even with smaller splits of annotated images, which can lead to competitive results with fewer labels. Additionally, assimilation enables us to quantify the propagated prior knowledge. This interpretability is another advantage that GCNs alone do not have.
	\begin{figure*}
	\begin{center}
		\includegraphics[width=0.99\textwidth]{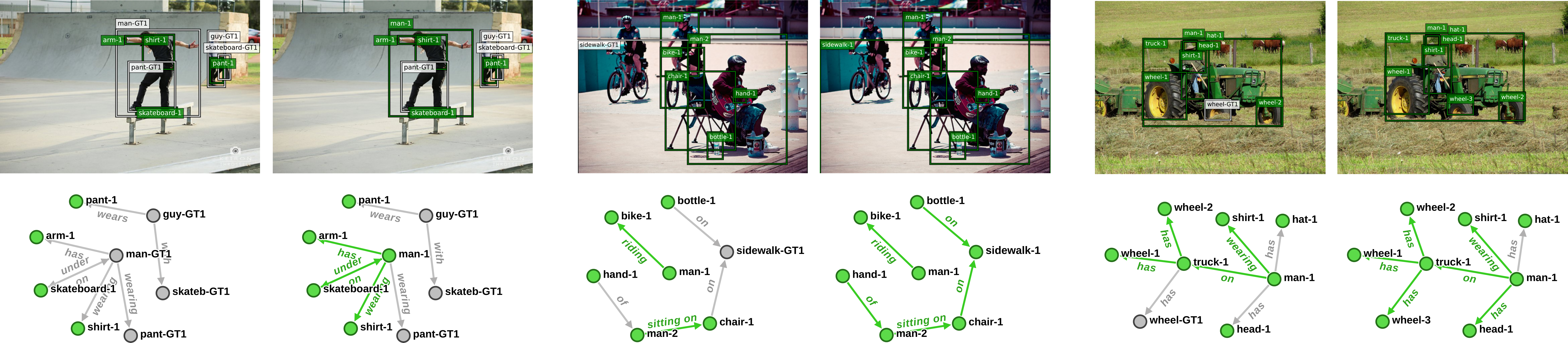}
	\end{center}
	\caption{Qualitative examples of improved scene graph classification results (Recall@50) through assimilations of our model. From left to right is after each assimilation. Green and gray colors indicate true positives and false negatives concluded by the model. For example consider the middle image, where the sidewalk was initially misclassified as a street. After seeing a biker in the image and a man sitting on a chair, a reasonable inference is that this should be a sidewalk.}\label{fig_qual}
\end{figure*}
	{\renewcommand{\arraystretch}{1.2}
		\begin{table*}
			\centering
			\small
			\scalebox{0.84}{
				\begin{tabular}{c|c|ccc|ccc|ccc}
					\hline
					\centering \multirow{2}{*}{Backbone} & 	\centering \multirow{2}{*}{Main Model} & \multicolumn{3}{c|}{SGCls R@100} & \multicolumn{3}{c|}{PredCls R@100}  & \multicolumn{3}{c}{Object Classification}\\
					& \multirow{2}{*} & $1{\%}$ & $10{\%}$ & $100{\%}$  & $1{\%}$ & $10{\%}$ & $100{\%}$  & $1{\%}$ & $10{\%}$ & $100{\%}$  \\
					\hline
					\hline
					
					\centering Supervised & \centering IC & $\phantom{0}1.84$ \tiny{$\pm0.26$} & $13.90$ \tiny{$\pm0.97$} & $33.6$ & $40.61$ \tiny{$\pm0.84$} & $52.51$ \tiny{$\pm1.19$} & $62.0$ & $14.38$ \tiny{$\pm 0.57$} & $38.45$ \tiny{$\pm 1.21$} & $64.2$\\
					\hdashline
					\centering Self-Supervised & \centering IC &$12.12$ \tiny{$\pm 0.47$} & $26.14$ \tiny{$\pm 0.77$} & $36.8$ & $48.10$ \tiny{$\pm 0.54$} & $58.14$ \tiny{$\pm 0.35$} & $63.4$ & $40.75$ \tiny{$\pm 0.48$} & $56.97$ \tiny{$\pm 0.76$} & $68.0$\\
					\centering \textbf{Self-Supervised} & \centering \textbf{IC + ICP} & \bm{$15.36$} \tiny{$\pm 0.38$} & \bm{$27.37$} \tiny{$\pm 0.47$} & \bm{$37.1$} & \bm{$65.68$} \tiny{$\pm 0.12$} & \bm{$65.42$} \tiny{$\pm 0.19$} & \bm{$65.7$} & \bm{$42.09$} \tiny{$\pm 0.65$} & \bm{$58.60$} \tiny{$\pm 0.56$} & \bm{$68.4$}\\
					
					\hline
			\end{tabular}}
			\caption{Comparison of R@100 for SGCls, PredCls and Object Classification tasks on smaller splits of the VG dataset.}
			\label{table_SSL}
		\end{table*}
	}
	
	{\renewcommand{\arraystretch}{1.2}
		\begin{table}
			\centering
			\small
			\scalebox{0.69}{
				\begin{tabular}{c|c|cc|cc|c}
					\hline
					\multirow{2}{*}{} & \centering \multirow{2}{*}{Method} & \multicolumn{2}{c|}{SGCls} & \multicolumn{2}{c|}{PredCls}  \\
					& & R@50 & R@100 &   R@50 & R@100 & Mean  \\
					\hline
					\hline
					\multirow{5}{*}{\rotatebox[origin=c]{90}{Unconstrained}} 
					& \centering IMP+ \citep{xu2017scene} & $43.4$ & $47.2$ & $75.2$ &  $83.6$ & $62.3$\\
					& \centering FREQ \citep{zellers2018neural}   & $40.5$ & $43.7$ &  $71.3$ & $81.$2 & $59.1$ \\
					& \centering SMN \citep{zellers2018neural} & $44.5$ & $47.7$  & $81.1$ & $88.3$& $65.4$\\
					& \centering KERN\citep{chen2019knowledge}   & $45.9$ & $49.0$  &  $81.9$ & $88.9$ &$66.4$  \\
					& \centering CCMT\citep{chen2019counterfactual}  & \bm{$48.6$} &   $52.0$ &  $83.2$ &  $90.1$  & $68.4$ \\
					\hdashline
					& \centering \textbf{Schemata}    & $48.5$ & \bm{$52.1$}  &  \bm{$83.8$} & \bm{$90.5$} &\bm{$68.7$}   \\
					\hline
					\multirow{6}{*}{\rotatebox[origin=c]{90}{Constrained}} 
					& \centering VRD \citep{lu2016visual}  &  $11.8$ & $14.1$ &  $27.9$  & $35.0$ & $22.2$  \\
					& \centering IMP \citep{xu2017scene}   &    $21.7$ & $24.4$ &   $44.8$ & $53.0$   & $36.9$\\
					& \centering IMP+ \citep{xu2017scene}  & $34.6$ & $35.4$ & $59.3$ & $61.3$ & $47.6$ \\
					& \centering FREQ \citep{zellers2018neural}  & $32.4$ & $34.0$ & $59.9$ & $64.1$ & $47.6$ \\
					& \centering SMN \citep{zellers2018neural}  & $35.8$ & $36.5$  & $65.2$ & $67.1$  & $51.1$\\
					& \centering KERN\citep{chen2019knowledge}  & $36.7$ & $37.4$ &  $65.8$ & $67.6$  & $51.8$  \\
					& \centering VCTree\citep{tang2019learning}  & $38.1$ &  $38.8$ &  $66.4$ &  $68.1$  & $52.8$ \\
					& \centering CCMT\citep{chen2019counterfactual}  & $39.0$ &   $39.8$ &  $66.4$ &  $68.1$  & $53.3$ \\
					\hdashline
					& \centering \textbf{Schemata}      &\bm{$39.1$} &\bm{$39.8$}               &\bm{$66.9$}&\bm{$68.4$}  &\bm{$53.5$} 
					\\
					& \centering Schemata - PKG      &--.- &--.-               &$48.9$&$54.2$  &--.-
					\\
					\hline
			\end{tabular}}
			\caption{Comparison of the R@50 and R@100, with and without graph constraints for SGCls and PredCls tasks on the VG dataset.}
			\label{table_R}
		\end{table}
	}
	\section{Evaluation}\label{sec_exp}
	{\renewcommand{\arraystretch}{1.2}
		\begin{table}
			\centering
			\small
			\scalebox{0.69}{
				\begin{tabular}{c|c|cc|cc|c}
					\hline
					\multirow{2}{*}{} & \centering \multirow{2}{*}{Method} & \multicolumn{2}{c|}{SGCls} & \multicolumn{2}{c|}{PredCls}  \\
					& & mR@50 & mR@100 &   mR@50 & mR@100 & Mean  \\
					\hline
					\hline
					\multirow{5}{*}{\rotatebox[origin=c]{90}{Unconstrained}} 
					& \centering IMP+ \citep{xu2017scene} &  $12.1$ & $16.9$ & $20.3$ & $28.9$ & $19.5$\\
					& \centering FREQ \citep{zellers2018neural} & $13.5$ & $19.6$ & $24.8$ & $37.3$ & $23.8$ \\
					& \centering SMN \citep{zellers2018neural} & $15.4$ & $20.6$ & $27.5$ & $37.9$ & $25.3$ \\
					& \centering KERN\citep{chen2019knowledge} & $19.8$ & $26.2$ & $36.3$ & $49.0$ & $32.8$\\
					\hdashline
					& \textbf{Schemata}      &\bm{$21.4$}        & \bm{$28.8$}    &\bm{$40.1$} &\bm{$54.9$} &\bm{$36.3$}\\
					\hline
					\multirow{6}{*}{\rotatebox[origin=c]{90}{Constrained}} 
					& \centering IMP \citep{xu2017scene} &  $\phantom{0}3.1$ & $\phantom{0}3.8$ & $\phantom{0}6.1$ & $\phantom{0}8.0$ & $\phantom{0}5.2$ \\
					& \centering IMP+ \citep{xu2017scene}  & $\phantom{0}5.8$ & $\phantom{0}6.0$ & $\phantom{0}9.8$ & $10.5$ & $\phantom{0}8.0$ \\
					& \centering FREQ \citep{zellers2018neural} & $\phantom{0}6.8$ & $\phantom{0}7.8$ & $13.3$ & $15.8$ & $10.9$\\
					
					& \centering SMN \citep{zellers2018neural}  & $\phantom{0}7.1$ & $\phantom{0}7.6$ & $13.3$ & $14.4$ & $10.6$\\
					& \centering KERN\citep{chen2019knowledge} & $\phantom{0}9.4$ & $10.0$ & $17.7$ & $19.2$ & $14.0$\\
					& \centering VCTree\citep{tang2019learning}  & $10.1$ &  $10.8$ &  $17.9$ &  $19.4$  & $14.5$ \\
					\hdashline
					& \textbf{Schemata}      &\bm{$10.1$}        & \bm{$10.9$}    &\bm{$19.1$} &\bm{$20.7$} &\bm{$15.2$} \\
					& \centering Schemata - PKG      &--.- &--.-               &$\phantom{0}8.2$&$\phantom{0}9.4$  &--.- \\
					\hline
			\end{tabular}}
			\caption{Comparison of the mR@50 and mR@100, with and without graph constraints for SGCls and PredCls tasks on the VG dataset.}
			\label{table_mR}
		\end{table}
	}
	\paragraph{Settings}
	We train our models on the common split of Visual Genome~\citep{krishna2017visual} dataset containing images labeled with their scene graphs~\citep{xu2017scene}. This split takes the most frequent 150 object and 50 predicate classes in total, with an average of 11.5 objects and 6.2 predicates in each image.
	We report the experimental results on the test set, under two standard classification settings of predicate classification \textbf{(PredCls)}: predicting predicate labels given a ground truth set of object boxes and object labels,
	and scene graph classification \textbf{(SGCls)}: predicting object and predicate labels, given the set of object boxes. 
	Another popular setting is the scene graph detection (SGDet), where the network should also detect the bounding boxes. Since the focus of our study is not on improving the object detector backbone and our improvements in SGDet are similar to the improvements in SGCls, we do not report them here. For those results, please refer to our official code repository.
	We report the results of these settings under \textit{constrained} and \textit{unconstrained} setups~\citep{yu2017visual}. In the unconstrained setup, we allow for multiple predicate labels, whereas in the constrained setup, we only take the top-1 predicted predicate label. 
	\paragraph{Metrics} 
	We use Recall@K (\textbf{R@K}) as the standard metric. R@K computes the mean prediction accuracy in each image given the top $K$ predictions.
	In VG, the distribution of labeled relations is highly imbalanced. Therefore, we additionally report Macro Recall~\citep{sharifzadeh2019improving,chen2019knowledge} (\textbf{mR@K}) to reflect the improvements in the long tail of the distribution. In this setting, the overall recall is computed by taking the mean over recall per predicate.
	
	\paragraph{Experiments}
	The goal of our experiments is \textbf{(A)} to study whether injecting prior knowledge into scene representations can improve the classification and \textbf{(B)} to study the commonsense knowledge that is captured in our model.
	In what follows, \textbf{\textit{backbone}} refers to VGG16/ResNet-50 that generates the \textit{SRG}, and \textbf{\textit{main model}} refers to part of the network that applies contextualization and assimilation. The backbone can be trained from a set of labeled images (in a supervised manner), unlabeled images (in a self-supervised manner), or a combination of the two. The main model can be trained from a set of labeled images (the \textit{IC} task), a prior knowledge base \textit{(ICP)} or a combination of the two. 
	Note that the memory consumption of the model is in the same order as Graph R-CNN and computing the extra attention heads can also be further optimized by estimating the softmax kernels~\citep{choromanski2020rethinking}.
	For (\textbf{A}), we conduct the following studies:
	\begin{enumerate}
		\item We train both the backbone and the main model from all the labeled images and for both tasks. We use the VGG-16 backbone as trained by~\citet{zellers2018neural}. This allows us to compare the results with the related works directly. We evaluate the classification accuracy for 8 assimilations (until the changes are not significant anymore). Table \ref{table_R} and \ref{table_mR} compare the performance of our model to the state-of-the-art. As shown, our model exceeds the others on average and under most settings. Figure \ref{fig_abl} shows our ablation study, indicating that the accuracy is improved after each assimilation.
		\item To qualitatively examine these results, we present some of the images and their scene graphs after two assimilations, in Figure \ref{fig_qual}. For example in the right image, while the wheel is almost fully occluded, we can still classify it once we classify other objects and employ commonsense (e.g., trucks have wheels). Another interesting example is the middle image, where the sidewalk is initially misclassified as a street. After seeing a biker in the image and a man sitting on a chair, a reasonable inference is that this should be a sidewalk! Similarly, in the left image, the man is facing away from the camera, and his pose makes it hard to classify him unless we utilize our prior knowledge about the arm, pants, shirt, and skateboard.
		\item Figure \ref{fig_perpred} shows the improvements per each predicate class. The results indicate that most improvements occur in under-represented classes. This means that we have achieved a generalization performance that is beyond the simple reflection of the dataset's statistical bias.
		\item To understand the importance of prior knowledge compared to having a large set of labeled images, we conduct the following study: we uniformly sample two splits with $1\%$ and $10\%$ of VG.
		The images in each split are considered as \textit{labeled}. We ignore the labels of the remaining images and consider them as unlabeled\footnote{Note that these splits are different from the recently proposed few-shot learning set by~\citet{chen2019scene}. In~\citep{chen2019scene}, the goal is to study the few-shot learning of \textit{predicates} only. However, we explore a more competitive setting, where only a fraction of both \textit{objects} and \textit{predicates} are labeled.}. Instead, we treat the set of ignored labels as a form of external/hand-crafted knowledge in the form of triples. For each split, we train the full model (\rom{1}) with a backbone that has been trained in a supervised fashion with the respective split and no pre-training, and the main model that has been trained for \textit{IC} (without commonsense) with the respective split, (\rom{2}) with a backbone that has been pre-trained on ImageNet~\citep{deng2009imagenet} and fine-tuned on the Visual Genome (in a self-supervised fashion with BYOL~\citep{grill2020bootstrap}) and fine-tuned on the respective split of the visual genome (in a supervised fashion) and the main model that has been trained for \textit{IC} with the respective split, and (\rom{3}) Similar to 2, except that we include the \textit{ICP} and train the main model by assimilating the entire prior knowledge base including the external triples. We discard their image-based features ($\mathbf{x}_i$) for the triples outside a split. Since BYOL is based on ResNet-50, for a fair comparison, we train all models in this experiment with ResNet-50 (including another model that we train with $100\%$ of the data). In the Scene Graph Classification community, the results are often reported under an arbitrary random seed, and previous works have not reported the summary statistics over several runs before. To allow for a fair comparison of our model to those works (on the $100\%$ set), we followed the same procedure in the study \textbf{A1}. However, to encourage a statistically more stable comparison of future models in this experiment, we report the summary statistics (arithmetic mean and standard deviation) over five random fractions ($1\%$ and $10\%$) of VG training set\footnote{The splits will be made publicly available.}. As shown in Table \ref{table_SSL}, utilizing prior knowledge allows to achieve almost the same predicate prediction accuracy with $1\%$ of the data only. Also, we largely improve object classification and scene graph classification.
	\end{enumerate}
	\begin{figure}
		\begin{center}
			\includegraphics[width=0.47\textwidth]{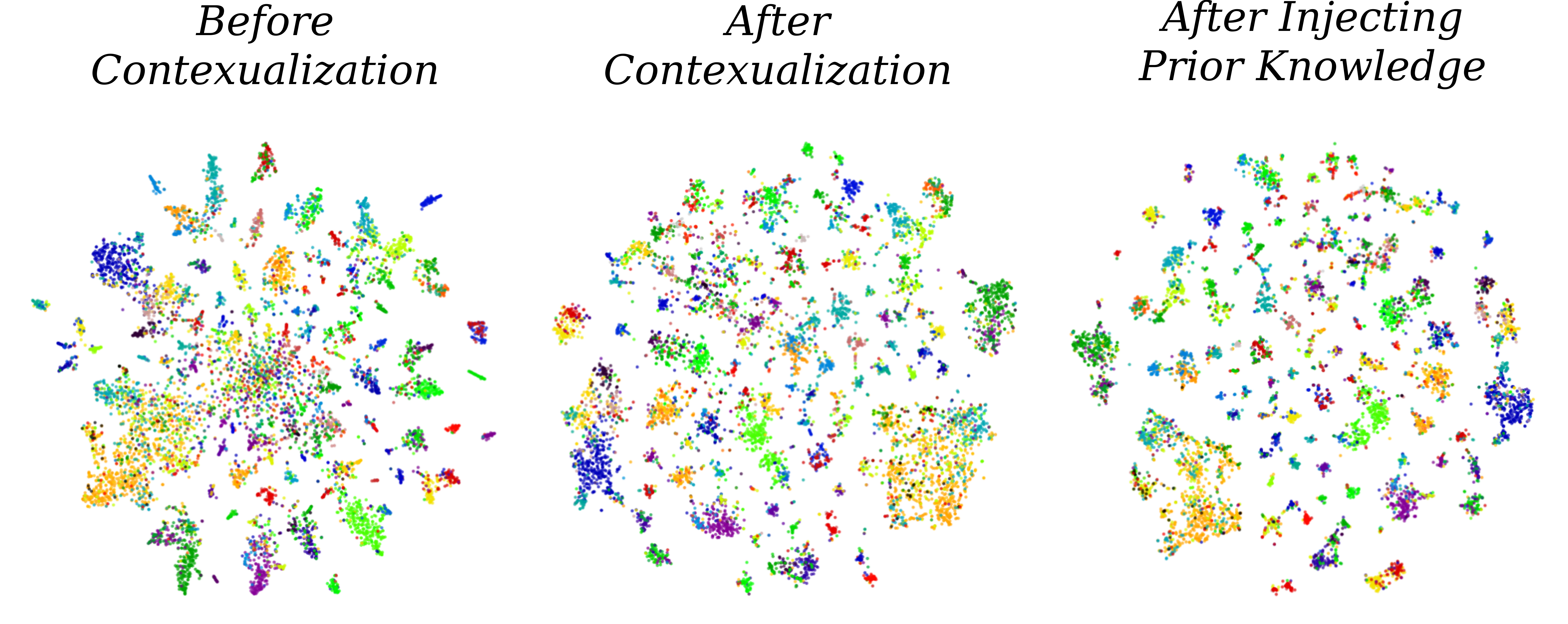}
		\end{center}
		\caption{t-SNE visualization of object representations before contextualization, after contextualization and after injecting prior knowledge.}\label{fig_tsne}
	\end{figure} 
	
	For \textbf{B} we consider the following studies: 
	\begin{enumerate}
		\item We visualize the semantic affinity of \textit{schema representations} by employing t-SNE~\citep{maaten2008visualizing}. As we can see in Figure \ref{fig_tsne_schema}, the schema representations of entities that are \textit{visually} or \textit{relationally} similar are the closest to each other.
		\item  We inspect the semantic affinity of \textit{object representations} by employing t-SNE (\rom{1}) before contextualization, (\rom{2}) after contextualization and (\rom{3}) after injecting prior knowledge. The results are represented in Figure \ref{fig_tsne}. Each color represents a different object class. This investigation confirms that object representations will get into more separable clusters after injecting prior knowledge.
		\item  Finally, we evaluate our model's accuracy in link prediction. The goal is to quantitatively evaluate our model's understanding of \textit{relational} commonsense, i.e., relational structure of the probabilistic knowledge graph. Similar to a KGE link prediction, we predict the predicate given \textit{head} and \textit{tail} of a relation. In other words, we feed our model with the schema of head and tail, together with a \textit{zero-vector} for the image-based representations. As we can see in Table \ref{table_mR} \& \ref{table_R}, in \textit{Schemata - PKG}, even if we do not provide any image-based information, our model can still \textit{guess} the expected predicates similar to a KGE model. While this guess is not as accurate as when we present it with an image, the accuracy is still remarkable.
	\end{enumerate}
	\begin{figure}
		\begin{center}
			\includegraphics[width=0.42\textwidth]{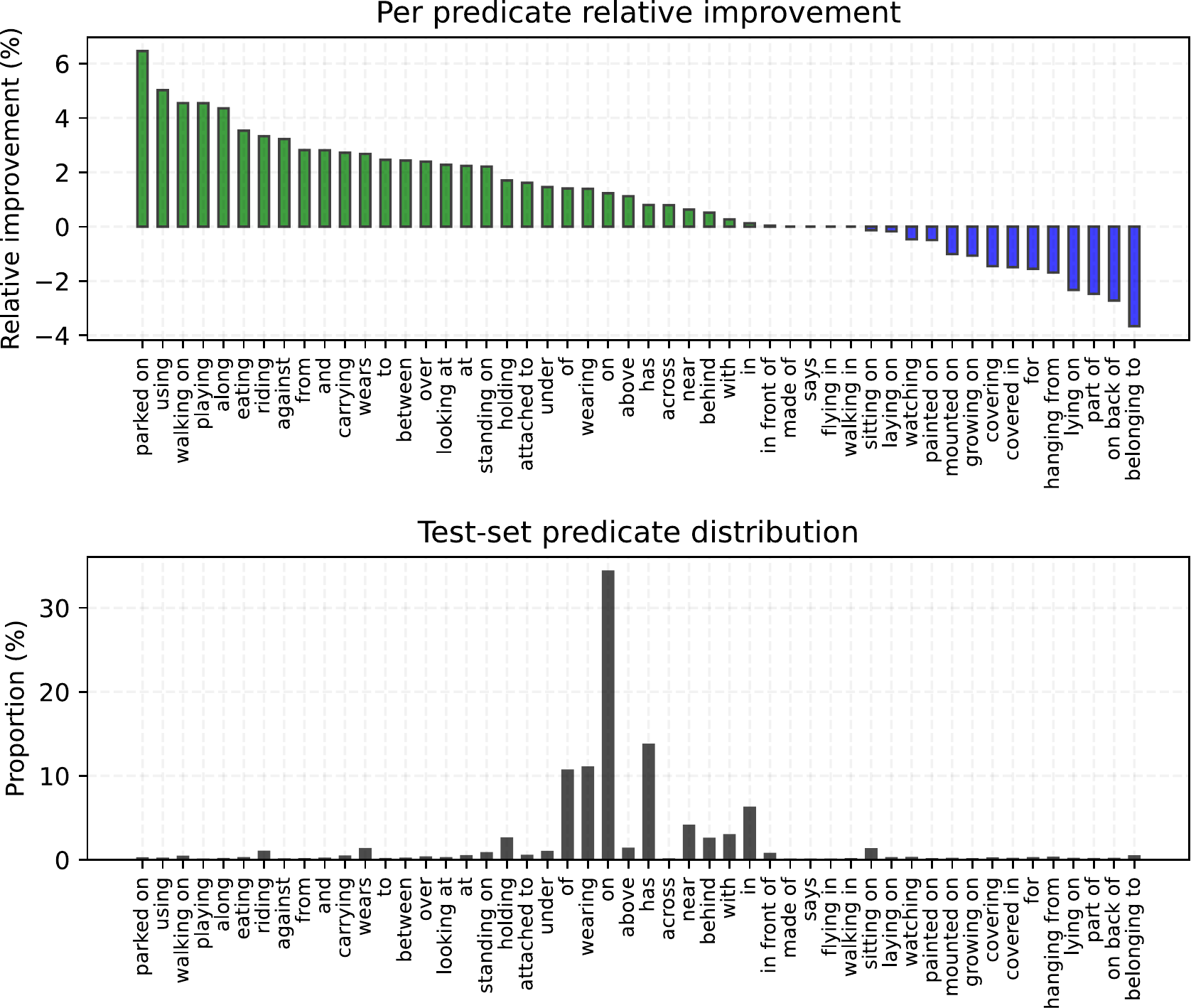}
		\end{center}
		\caption{The top plot shows the per predicate classification accuracy improvement, in SGCls R@100 with no graph constraints and from Assimiliation 1 to Assimilation 8. The bottom plot shows the distribution of sample proportion for the predicates in Visual Genome.}\label{fig_perpred}
	\end{figure}
	\section{Conclusion}\label{sec_conclusion}
	We discussed schemata as mental representations that enable compositionality and reasoning. To model schemata in a deep learning framework, we introduced them as representations that encode image-based and relational prior knowledge of objects and predicates in each class. By defining classification as an attention layer instead of a fully connected layer, we introduced an inductive bias that enabled the propagation of prior knowledge. 
	Our experiments on the Visual Genome dataset confirmed the effectiveness of assimilation through qualitative and quantitative measures. Our model achieved higher accuracy under most settings and could also accurately predict the commonsense knowledge. Additionally, we showed that our model could be fine-tuned from external sources of knowledge in the form of triples. When combined with pre-trained schemata in a self-supervised setting, this leads to a predicate prediction accuracy that is almost equal to the full model. Also, it gives significant improvements in the scene graph and object classification tasks. We hope that this work will open new research directions in utilizing commonsense to learn from little annotations.
	\section{Acknowledgments}
	We would like to thank Max Berrendorf, Dario Konopatzki, Shaya Akbarinejad, Lisa Machata, Shabnam Sadegh, and the anonymous reviewers for the fruitful discussions and helpful feedback on the manuscript. 
	\bibliography{aaai21}

\begin{thebibliography}{53}
\providecommand{\natexlab}[1]{#1}
\providecommand{\url}[1]{\texttt{#1}}
\providecommand{\urlprefix}{URL }
\expandafter\ifx\csname urlstyle\endcsname\relax
  \providecommand{\doi}[1]{doi:\discretionary{}{}{}#1}\else
  \providecommand{\doi}{doi:\discretionary{}{}{}\begingroup
  \urlstyle{rm}\Url}\fi

\bibitem[{Arbib(1992)}]{arbib1992schema}
Arbib, M.~A. 1992.
\newblock Schema theory.
\newblock \emph{The encyclopedia of artificial intelligence} 2: 1427--1443.

\bibitem[{Ba, Kiros, and Hinton(2016)}]{ba2016layer}
Ba, J.~L.; Kiros, J.~R.; and Hinton, G.~E. 2016.
\newblock Layer normalization.
\newblock \emph{arXiv preprint arXiv:1607.06450} .

\bibitem[{Baier, Ma, and Tresp(2017)}]{baier2017improving}
Baier, S.; Ma, Y.; and Tresp, V. 2017.
\newblock Improving visual relationship detection using semantic modeling of
  scene descriptions.
\newblock In \emph{International Semantic Web Conference}, 53--68. Springer.

\bibitem[{Baier, Ma, and Tresp(2018)}]{baier2018improving}
Baier, S.; Ma, Y.; and Tresp, V. 2018.
\newblock Improving information extraction from images with learned semantic
  models.
\newblock In \emph{Proceedings of the 27th International Joint Conference on
  Artificial Intelligence}, 5214--5218. AAAI Press.

\bibitem[{Battaglia et~al.(2018)Battaglia, Hamrick, Bapst, Sanchez-Gonzalez,
  Zambaldi, Malinowski, Tacchetti, Raposo, Santoro, Faulkner
  et~al.}]{battaglia2018relational}
Battaglia, P.~W.; Hamrick, J.~B.; Bapst, V.; Sanchez-Gonzalez, A.; Zambaldi,
  V.; Malinowski, M.; Tacchetti, A.; Raposo, D.; Santoro, A.; Faulkner, R.;
  et~al. 2018.
\newblock Relational inductive biases, deep learning, and graph networks.
\newblock \emph{arXiv preprint arXiv:1806.01261} .

\bibitem[{Bengio et~al.(2015)Bengio, Vinyals, Jaitly, and
  Shazeer}]{bengio2015scheduled}
Bengio, S.; Vinyals, O.; Jaitly, N.; and Shazeer, N. 2015.
\newblock Scheduled sampling for sequence prediction with recurrent neural
  networks.
\newblock In \emph{Advances in Neural Information Processing Systems},
  1171--1179.

\bibitem[{Bengio(2017)}]{bengio2017consciousness}
Bengio, Y. 2017.
\newblock The consciousness prior.
\newblock \emph{arXiv preprint arXiv:1709.08568} .

\bibitem[{Bordes et~al.(2013)Bordes, Usunier, Garcia-Duran, Weston, and
  Yakhnenko}]{bordes2013translating}
Bordes, A.; Usunier, N.; Garcia-Duran, A.; Weston, J.; and Yakhnenko, O. 2013.
\newblock Translating embeddings for modeling multi-relational data.
\newblock In \emph{Advances in neural information processing systems},
  2787--2795.

\bibitem[{Caruana(1997)}]{caruana1997multitask}
Caruana, R. 1997.
\newblock Multitask learning.
\newblock \emph{Machine learning} 28(1): 41--75.

\bibitem[{Chen et~al.(2019{\natexlab{a}})Chen, Zhang, Xiao, He, Pu, and
  Chang}]{chen2019counterfactual}
Chen, L.; Zhang, H.; Xiao, J.; He, X.; Pu, S.; and Chang, S.-F.
  2019{\natexlab{a}}.
\newblock Counterfactual critic multi-agent training for scene graph
  generation.
\newblock In \emph{Proceedings of the IEEE International Conference on Computer
  Vision}, 4613--4623.

\bibitem[{Chen et~al.(2019{\natexlab{b}})Chen, Xu, Hui, Wu, and
  Lin}]{chen2019learning}
Chen, T.; Xu, M.; Hui, X.; Wu, H.; and Lin, L. 2019{\natexlab{b}}.
\newblock Learning semantic-specific graph representation for multi-label image
  recognition.
\newblock In \emph{Proceedings of the IEEE International Conference on Computer
  Vision}, 522--531.

\bibitem[{Chen et~al.(2019{\natexlab{c}})Chen, Yu, Chen, and
  Lin}]{chen2019knowledge}
Chen, T.; Yu, W.; Chen, R.; and Lin, L. 2019{\natexlab{c}}.
\newblock Knowledge-embedded routing network for scene graph generation.
\newblock In \emph{Proceedings of the IEEE Conference on Computer Vision and
  Pattern Recognition}, 6163--6171.

\bibitem[{Chen et~al.(2019{\natexlab{d}})Chen, Varma, Krishna, Bernstein, Re,
  and Fei-Fei}]{chen2019scene}
Chen, V.~S.; Varma, P.; Krishna, R.; Bernstein, M.; Re, C.; and Fei-Fei, L.
  2019{\natexlab{d}}.
\newblock Scene graph prediction with limited labels.
\newblock In \emph{Proceedings of the IEEE International Conference on Computer
  Vision}, 2580--2590.

\bibitem[{Choromanski et~al.(2020)Choromanski, Likhosherstov, Dohan, Song,
  Gane, Sarlos, Hawkins, Davis, Mohiuddin, Kaiser
  et~al.}]{choromanski2020rethinking}
Choromanski, K.; Likhosherstov, V.; Dohan, D.; Song, X.; Gane, A.; Sarlos, T.;
  Hawkins, P.; Davis, J.; Mohiuddin, A.; Kaiser, L.; et~al. 2020.
\newblock Rethinking attention with performers.
\newblock \emph{arXiv preprint arXiv:2009.14794} .

\bibitem[{Cs{\'a}ji et~al.(2001)}]{csaji2001approximation}
Cs{\'a}ji, B.~C.; et~al. 2001.
\newblock Approximation with artificial neural networks.
\newblock \emph{Faculty of Sciences, Etvs Lornd University, Hungary} 24(48): 7.

\bibitem[{Deng et~al.(2014)Deng, Ding, Jia, Frome, Murphy, Bengio, Li, Neven,
  and Adam}]{deng2014large}
Deng, J.; Ding, N.; Jia, Y.; Frome, A.; Murphy, K.; Bengio, S.; Li, Y.; Neven,
  H.; and Adam, H. 2014.
\newblock Large-scale object classification using label relation graphs.
\newblock In \emph{European conference on computer vision}, 48--64. Springer.

\bibitem[{Deng et~al.(2009)Deng, Dong, Socher, Li, Li, and
  Fei-Fei}]{deng2009imagenet}
Deng, J.; Dong, W.; Socher, R.; Li, L.-J.; Li, K.; and Fei-Fei, L. 2009.
\newblock Imagenet: A large-scale hierarchical image database.
\newblock In \emph{2009 IEEE conference on computer vision and pattern
  recognition}, 248--255. Ieee.

\bibitem[{Fodor, Pylyshyn et~al.(1988)}]{fodor1988connectionism}
Fodor, J.~A.; Pylyshyn, Z.~W.; et~al. 1988.
\newblock Connectionism and cognitive architecture: A critical analysis.
\newblock \emph{Cognition} 28(1-2): 3--71.

\bibitem[{Glorot and Bengio(2010)}]{glorot2010understanding}
Glorot, X.; and Bengio, Y. 2010.
\newblock Understanding the difficulty of training deep feedforward neural
  networks.
\newblock In \emph{Proceedings of the thirteenth international conference on
  artificial intelligence and statistics}, 249--256.

\bibitem[{Goyal et~al.(2020)Goyal, Lamb, Gampa, Beaudoin, Levine, Blundell,
  Bengio, and Mozer}]{goyal2020object}
Goyal, A.; Lamb, A.; Gampa, P.; Beaudoin, P.; Levine, S.; Blundell, C.; Bengio,
  Y.; and Mozer, M. 2020.
\newblock Object Files and Schemata: Factorizing Declarative and Procedural
  Knowledge in Dynamical Systems.
\newblock \emph{arXiv preprint arXiv:2006.16225} .

\bibitem[{Grill et~al.(2020)Grill, Strub, Altch{\'e}, Tallec, Richemond,
  Buchatskaya, Doersch, Pires, Guo, Azar et~al.}]{grill2020bootstrap}
Grill, J.-B.; Strub, F.; Altch{\'e}, F.; Tallec, C.; Richemond, P.~H.;
  Buchatskaya, E.; Doersch, C.; Pires, B.~A.; Guo, Z.~D.; Azar, M.~G.; et~al.
  2020.
\newblock Bootstrap your own latent: A new approach to self-supervised
  learning.
\newblock \emph{arXiv preprint arXiv:2006.07733} .

\bibitem[{He et~al.(2016)He, Zhang, Ren, and Sun}]{he2016deep}
He, K.; Zhang, X.; Ren, S.; and Sun, J. 2016.
\newblock Deep residual learning for image recognition.
\newblock In \emph{Proceedings of the IEEE conference on computer vision and
  pattern recognition}, 770--778.

\bibitem[{Hochreiter and Schmidhuber(1997)}]{hochreiter1997long}
Hochreiter, S.; and Schmidhuber, J. 1997.
\newblock Long short-term memory.
\newblock \emph{Neural computation} 9(8): 1735--1780.

\bibitem[{Hou et~al.(2019)Hou, Wu, Qi, Zhao, Luo, and Jia}]{hou2019relational}
Hou, J.; Wu, X.; Qi, Y.; Zhao, W.; Luo, J.; and Jia, Y. 2019.
\newblock Relational Reasoning using Prior Knowledge for Visual Captioning.
\newblock \emph{arXiv preprint arXiv:1906.01290} .

\bibitem[{Hu et~al.(2017)Hu, Deng, Zhou, Sha, and Mori}]{hu2017labelbank}
Hu, H.; Deng, Z.; Zhou, G.-T.; Sha, F.; and Mori, G. 2017.
\newblock Labelbank: Revisiting global perspectives for semantic segmentation.
\newblock \emph{arXiv preprint arXiv:1703.09891} .

\bibitem[{Hu et~al.(2016)Hu, Zhou, Deng, Liao, and Mori}]{hu2016learning}
Hu, H.; Zhou, G.-T.; Deng, Z.; Liao, Z.; and Mori, G. 2016.
\newblock Learning structured inference neural networks with label relations.
\newblock In \emph{Proceedings of the IEEE Conference on Computer Vision and
  Pattern Recognition}, 2960--2968.

\bibitem[{Kansky et~al.(2017)Kansky, Silver, M{\'e}ly, Eldawy,
  L{\'a}zaro-Gredilla, Lou, Dorfman, Sidor, Phoenix, and
  George}]{kansky2017schema}
Kansky, K.; Silver, T.; M{\'e}ly, D.~A.; Eldawy, M.; L{\'a}zaro-Gredilla, M.;
  Lou, X.; Dorfman, N.; Sidor, S.; Phoenix, S.; and George, D. 2017.
\newblock Schema networks: Zero-shot transfer with a generative causal model of
  intuitive physics.
\newblock \emph{arXiv preprint arXiv:1706.04317} .

\bibitem[{Kant(1787)}]{kant2011kritik}
Kant, I. 1787.
\newblock \emph{Kritik der reinen Vernunft:[Hauptband]}.
\newblock Walter de Gruyter.

\bibitem[{Kingma and Ba(2014)}]{kingma2014adam}
Kingma, D.~P.; and Ba, J. 2014.
\newblock Adam: A method for stochastic optimization.
\newblock \emph{arXiv preprint arXiv:1412.6980} .

\bibitem[{Kipf and Welling(2016)}]{kipf2016semi}
Kipf, T.~N.; and Welling, M. 2016.
\newblock Semi-supervised classification with graph convolutional networks.
\newblock \emph{arXiv preprint arXiv:1609.02907} .

\bibitem[{Koncel-Kedziorski et~al.(2019)Koncel-Kedziorski, Bekal, Luan, Lapata,
  and Hajishirzi}]{koncel2019text}
Koncel-Kedziorski, R.; Bekal, D.; Luan, Y.; Lapata, M.; and Hajishirzi, H.
  2019.
\newblock Text generation from knowledge graphs with graph transformers.
\newblock \emph{arXiv preprint arXiv:1904.02342} .

\bibitem[{Krishna et~al.(2017)Krishna, Zhu, Groth, Johnson, Hata, Kravitz,
  Chen, Kalantidis, Li, Shamma et~al.}]{krishna2017visual}
Krishna, R.; Zhu, Y.; Groth, O.; Johnson, J.; Hata, K.; Kravitz, J.; Chen, S.;
  Kalantidis, Y.; Li, L.-J.; Shamma, D.~A.; et~al. 2017.
\newblock Visual genome: Connecting language and vision using crowdsourced
  dense image annotations.
\newblock \emph{International Journal of Computer Vision} 123(1): 32--73.

\bibitem[{Lu et~al.(2016)Lu, Krishna, Bernstein, and Fei-Fei}]{lu2016visual}
Lu, C.; Krishna, R.; Bernstein, M.; and Fei-Fei, L. 2016.
\newblock Visual relationship detection with language priors.
\newblock In \emph{European Conference on Computer Vision}, 852--869. Springer.

\bibitem[{Maaten and Hinton(2008)}]{maaten2008visualizing}
Maaten, L. v.~d.; and Hinton, G. 2008.
\newblock Visualizing data using t-SNE.
\newblock \emph{Journal of machine learning research} 9(Nov): 2579--2605.

\bibitem[{Marcus(2018)}]{marcus2018deep}
Marcus, G. 2018.
\newblock Deep learning: A critical appraisal.
\newblock \emph{arXiv preprint arXiv:1801.00631} .

\bibitem[{Montufar et~al.(2014)Montufar, Pascanu, Cho, and
  Bengio}]{montufar2014number}
Montufar, G.~F.; Pascanu, R.; Cho, K.; and Bengio, Y. 2014.
\newblock On the number of linear regions of deep neural networks.
\newblock In \emph{Advances in neural information processing systems},
  2924--2932.

\bibitem[{Nickel et~al.(2016)Nickel, Murphy, Tresp, and
  Gabrilovich}]{nickel2016review}
Nickel, M.; Murphy, K.; Tresp, V.; and Gabrilovich, E. 2016.
\newblock A review of relational machine learning for knowledge graphs.
\newblock \emph{Proceedings of the IEEE} 104(1): 11--33.

\bibitem[{Nickel, Tresp, and Kriegel(2011)}]{nickel2011three}
Nickel, M.; Tresp, V.; and Kriegel, H.-P. 2011.
\newblock A three-way model for collective learning on multi-relational data.
\newblock In \emph{Icml}, volume~11, 809--816.

\bibitem[{Piaget(1923)}]{piaget1923langage}
Piaget, J. 1923.
\newblock \emph{Langage et pensée chez l'enfant}.
\newblock Delachaux et Niestlé.

\bibitem[{Ren et~al.(2015)Ren, He, Girshick, and Sun}]{ren2015faster}
Ren, S.; He, K.; Girshick, R.; and Sun, J. 2015.
\newblock Faster r-cnn: Towards real-time object detection with region proposal
  networks.
\newblock In \emph{Advances in neural information processing systems}, 91--99.

\bibitem[{Santoro et~al.(2017)Santoro, Raposo, Barrett, Malinowski, Pascanu,
  Battaglia, and Lillicrap}]{santoro2017simple}
Santoro, A.; Raposo, D.; Barrett, D.~G.; Malinowski, M.; Pascanu, R.;
  Battaglia, P.; and Lillicrap, T. 2017.
\newblock A simple neural network module for relational reasoning.
\newblock In \emph{Advances in neural information processing systems},
  4967--4976.

\bibitem[{Sharifzadeh et~al.(2019)Sharifzadeh, Moayed~Baharlou, Berrendorf,
  Koner, and Tresp}]{sharifzadeh2019improving}
Sharifzadeh, S.; Moayed~Baharlou, S.; Berrendorf, M.; Koner, R.; and Tresp, V.
  2019.
\newblock Improving Visual Relation Detection using Depth Maps.
\newblock \emph{arXiv preprint arXiv:1905.00966} .

\bibitem[{Simonyan and Zisserman(2014)}]{simonyan2014very}
Simonyan, K.; and Zisserman, A. 2014.
\newblock Very deep convolutional networks for large-scale image recognition.
\newblock \emph{arXiv preprint arXiv:1409.1556} .

\bibitem[{Tang et~al.(2019)Tang, Zhang, Wu, Luo, and Liu}]{tang2019learning}
Tang, K.; Zhang, H.; Wu, B.; Luo, W.; and Liu, W. 2019.
\newblock Learning to compose dynamic tree structures for visual contexts.
\newblock In \emph{Proceedings of the IEEE Conference on Computer Vision and
  Pattern Recognition}, 6619--6628.

\bibitem[{Tresp, Sharifzadeh, and Konopatzki(2019)}]{tresp2019model}
Tresp, V.; Sharifzadeh, S.; and Konopatzki, D. 2019.
\newblock A Model for Perception and Memory.

\bibitem[{Tresp et~al.(2020)Tresp, Sharifzadeh, Konopatzki, and
  Ma}]{tresp2020tensor}
Tresp, V.; Sharifzadeh, S.; Konopatzki, D.; and Ma, Y. 2020.
\newblock The Tensor Brain: Semantic Decoding for Perception and Memory.
\newblock \emph{arXiv preprint arXiv:2001.11027} .

\bibitem[{Vaswani et~al.(2017)Vaswani, Shazeer, Parmar, Uszkoreit, Jones,
  Gomez, Kaiser, and Polosukhin}]{vaswani2017attention}
Vaswani, A.; Shazeer, N.; Parmar, N.; Uszkoreit, J.; Jones, L.; Gomez, A.~N.;
  Kaiser, {\L}.; and Polosukhin, I. 2017.
\newblock Attention is all you need.
\newblock In \emph{Advances in neural information processing systems},
  5998--6008.

\bibitem[{Wu, Lenz, and Saxena(2014)}]{wu2014hierarchical}
Wu, C.; Lenz, I.; and Saxena, A. 2014.
\newblock Hierarchical Semantic Labeling for Task-Relevant RGB-D Perception.
\newblock In \emph{Robotics: Science and systems}.

\bibitem[{Xu et~al.(2017)Xu, Zhu, Choy, and Fei-Fei}]{xu2017scene}
Xu, D.; Zhu, Y.; Choy, C.~B.; and Fei-Fei, L. 2017.
\newblock Scene graph generation by iterative message passing.
\newblock In \emph{Proceedings of the IEEE Conference on Computer Vision and
  Pattern Recognition}, 5410--5419.

\bibitem[{Yang et~al.(2018)Yang, Lu, Lee, Batra, and Parikh}]{yang2018graph}
Yang, J.; Lu, J.; Lee, S.; Batra, D.; and Parikh, D. 2018.
\newblock Graph r-cnn for scene graph generation.
\newblock In \emph{Proceedings of the European Conference on Computer Vision
  (ECCV)}, 670--685.

\bibitem[{Yu et~al.(2017)Yu, Li, Morariu, and Davis}]{yu2017visual}
Yu, R.; Li, A.; Morariu, V.~I.; and Davis, L.~S. 2017.
\newblock Visual relationship detection with internal and external linguistic
  knowledge distillation.
\newblock In \emph{IEEE International Conference on Computer Vision (ICCV)}.

\bibitem[{Zellers et~al.(2018)Zellers, Yatskar, Thomson, and
  Choi}]{zellers2018neural}
Zellers, R.; Yatskar, M.; Thomson, S.; and Choi, Y. 2018.
\newblock Neural motifs: Scene graph parsing with global context.
\newblock In \emph{Proceedings of the IEEE Conference on Computer Vision and
  Pattern Recognition}, 5831--5840.

\bibitem[{Zhang et~al.(2017)Zhang, Kyaw, Chang, and Chua}]{zhang2017visual}
Zhang, H.; Kyaw, Z.; Chang, S.; and Chua, T. 2017.
\newblock Visual Translation Embedding Network for Visual Relation Detection.
\newblock In \emph{2017 {IEEE} Conference on Computer Vision and Pattern
  Recognition, {CVPR} 2017, Honolulu, HI, USA, July 21-26, 2017}, 3107--3115.
  {IEEE} Computer Society.
\newblock ISBN 978-1-5386-0457-1.
\newblock \doi{10.1109/CVPR.2017.331}.
\newblock \urlprefix\url{https://doi.org/10.1109/CVPR.2017.331}.

\end{thebibliography}
		\cleardoublepage
		\appendix


		\section{Architectural Details}
		As discussed before, for the experiments A1-3, similar to ~\cite{zellers2018neural} and following works, we use Faster R-CNN~\citep{ren2015faster} with VGG-16~\citep{simonyan2014very} and for the A4, we use ResNet-50.
		After extracting the image embeddings from the penultimate fully connected (fc) layer of the backbones, we feed them to a fc-layer with 512 neurons and a Leaky ReLU with a slope of 0.2, together with a dropout rate of 0.8. This gives us initial object node embeddings. We apply a fc-layer with 512 neurons and Leaky ReLU with a slope 0.2 and dropout rate of 0.1, to the extracted spatial vector $\mathbf{t}$ to initialize predicate embeddings. For the contextualization, we take four graph transformer layers  of 5 heads each. $f$ and  $g$ have 2048 and 512 neurons in each layer. We initial the layers with using Glorot weights~\citep{glorot2010understanding}. 
		\section{Training with Full Data (A1-3)}
		We trained our full model for SGCls setting with Adam optimizer~\citep{kingma2014adam} and a learning rate of $10^{-5}$ for 24 epochs with a batch size of 14. We fine-tuned this model for PredCls with 10 epochs, batchsize of 24, and SGD optimizer with learning rate of $10^{-4}$. We used an NVIDIA RTX-2080ti GPU for training.
		\section{Training with Splits of Data (A4)}
		\subsection{Training Details of the Backbones}
		\subsubsection{Supervised:} We train the supervised backbones on the corresponding split of visual genome training set with the Adam optimizer~\citep{kingma2014adam} and a learning rate of $10^{-5}$ for 20 epochs with a batch size of 6.
		\subsubsection{Self-supervised:} We train the self-supervised backbones with the BYOL~\citep{grill2020bootstrap} approach. We fine-tune the pre-trained \textit{self-supervised} weights over ImageNet~\citep{deng2009imagenet}, on the entire training set of Visual Genome images in a self-supervised manner with no labels, for 3 epochs with a batch-size of 6, SGD optimizer with a learning rate of $6 \times 10^{-5}$, momentum of 0.9 and weight decay of $4 \times 10^{-4}$. Similar to BYOL, we use a MLP hidden size of 512 and a projection layer size of 128 neurons. Then for each corresponding split, we fine-tune the weights in a supervised manner with the Adam optimizer and a learning rate of $10^{-5}$ for 4 epochs with a batch size of 6.
		\subsection{Training Details of the Main Models}
		\subsubsection{Supervised + IC}
		We train the main model with the Adam optimizer and a learning rate of $10^{-5}$ and a batch size of 22, with 5 epochs for the $1\%$ and 11 epochs for $10\%$.
		\subsubsection{Self-supervised + IC}
		We train the main model with the Adam optimizer and a learning rate of $10^{-5}$ and a batch size of 22, with 6 epochs for the $1\%$ and 11 epochs for $10\%$.
		\subsubsection{Self-supervised + \textbf{IC + ICP}}
		We train the main model with the Adam optimizer and a learning rate of $10^{-5}$ with a batch size of 16, with 6 epochs for the $1\%$ and 11 epochs for $10\%$. We train these models with 2 assimilations. 	
		\section{Additional Qualitative Results}
		In Figure \ref{fig_qual1} and \ref{fig_qual2}, we present additional qualitative results on the Assimilation Study (Figure 4, Table 1 and Table 2 in the main paper). In each assimilation either object or predicate classification are improved.
		\section{Per-predicate Improvement}
\begin{figure}
			\begin{center}
				\includegraphics[width=0.45\textwidth]{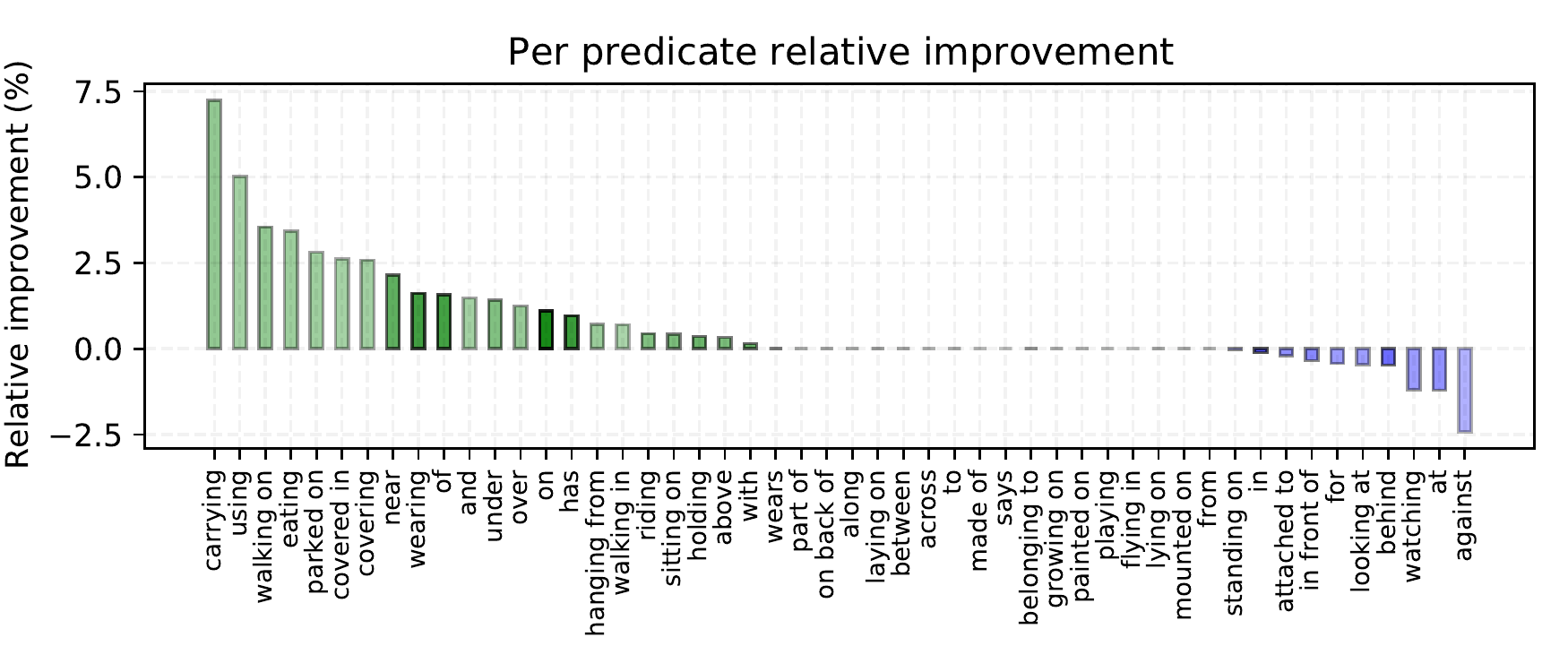}
			\end{center}
			\caption{ R@50 - with graph constraints}\label{fig_perpred1}
		\end{figure}

\begin{figure}
			\begin{center}
				\includegraphics[width=0.45\textwidth]{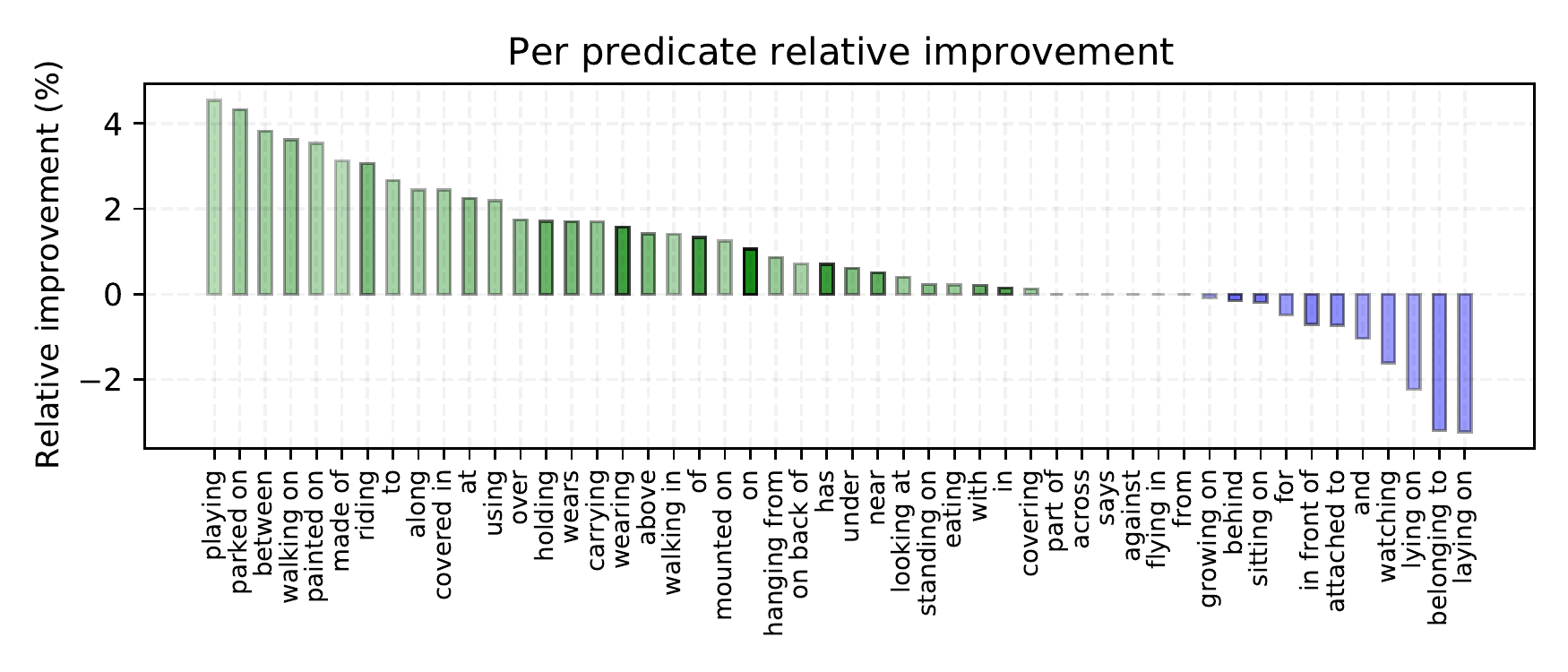}
			\end{center}
			\caption{R@50 - with no graph constraints}\label{fig_perpred2}
		\end{figure}

\begin{figure}[!t]
			\begin{center}
				\includegraphics[width=0.45\textwidth]{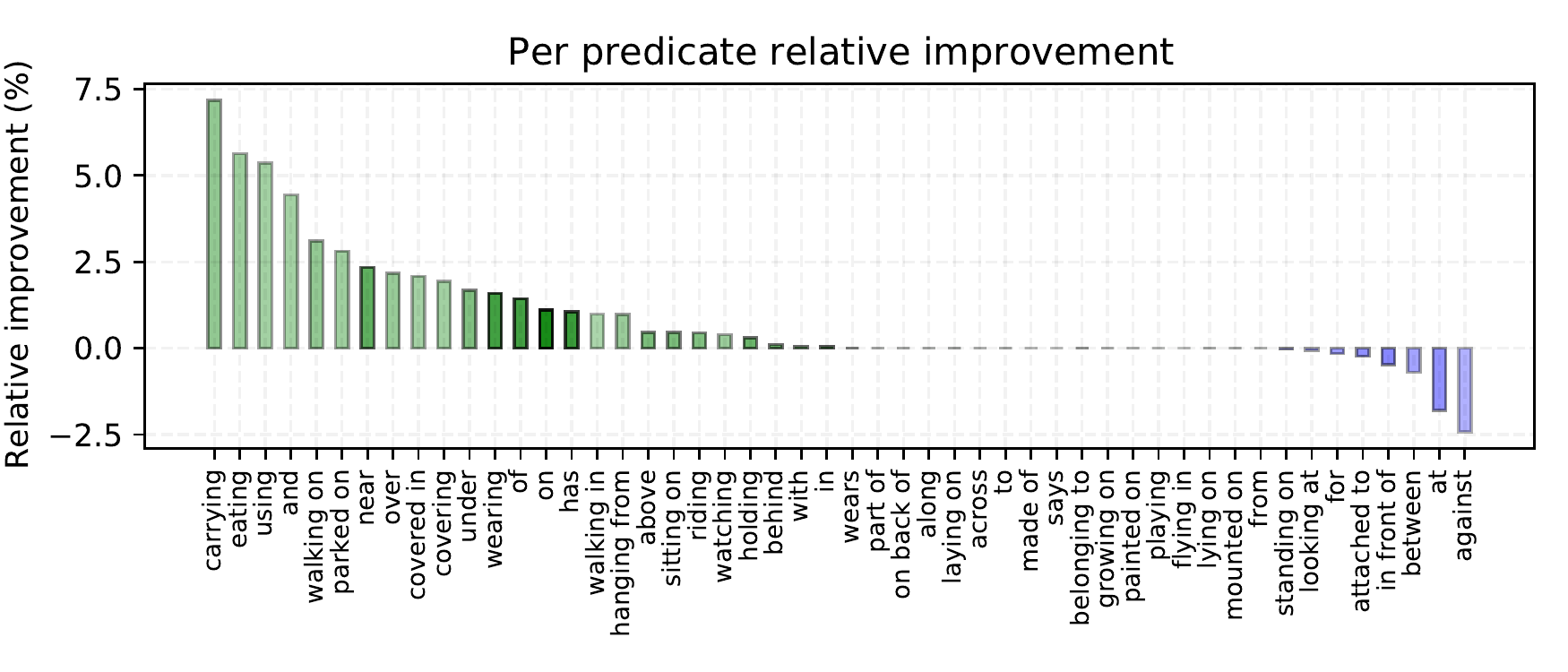}
			\end{center}
			\caption{R@100 - with graph constraints}\label{fig_perpred3}
		\end{figure}

\begin{figure*}
			\begin{center}
				\includegraphics[width=0.7\textwidth]{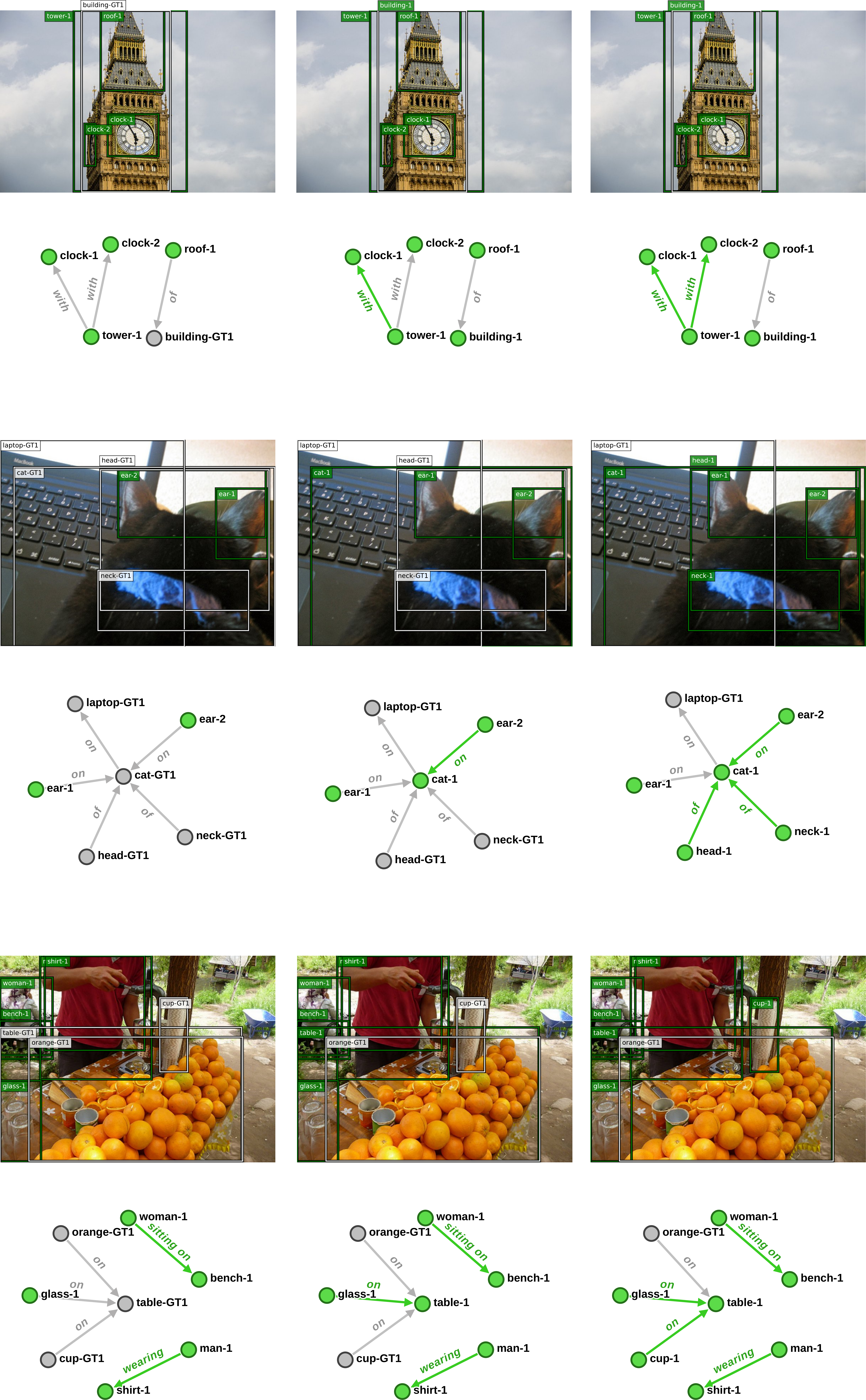}
			\end{center}
			\caption{Qualitative examples of improved scene graph classification results (R@50 - with graph constraints) through assimilations of our model. From left to right is after each assimilation. Green and gray colors indicate true positives and false negatives concluded by the model.}\label{fig_qual1}
		\end{figure*}

\begin{figure*}
			\begin{center}
				\includegraphics[width=0.92\textwidth]{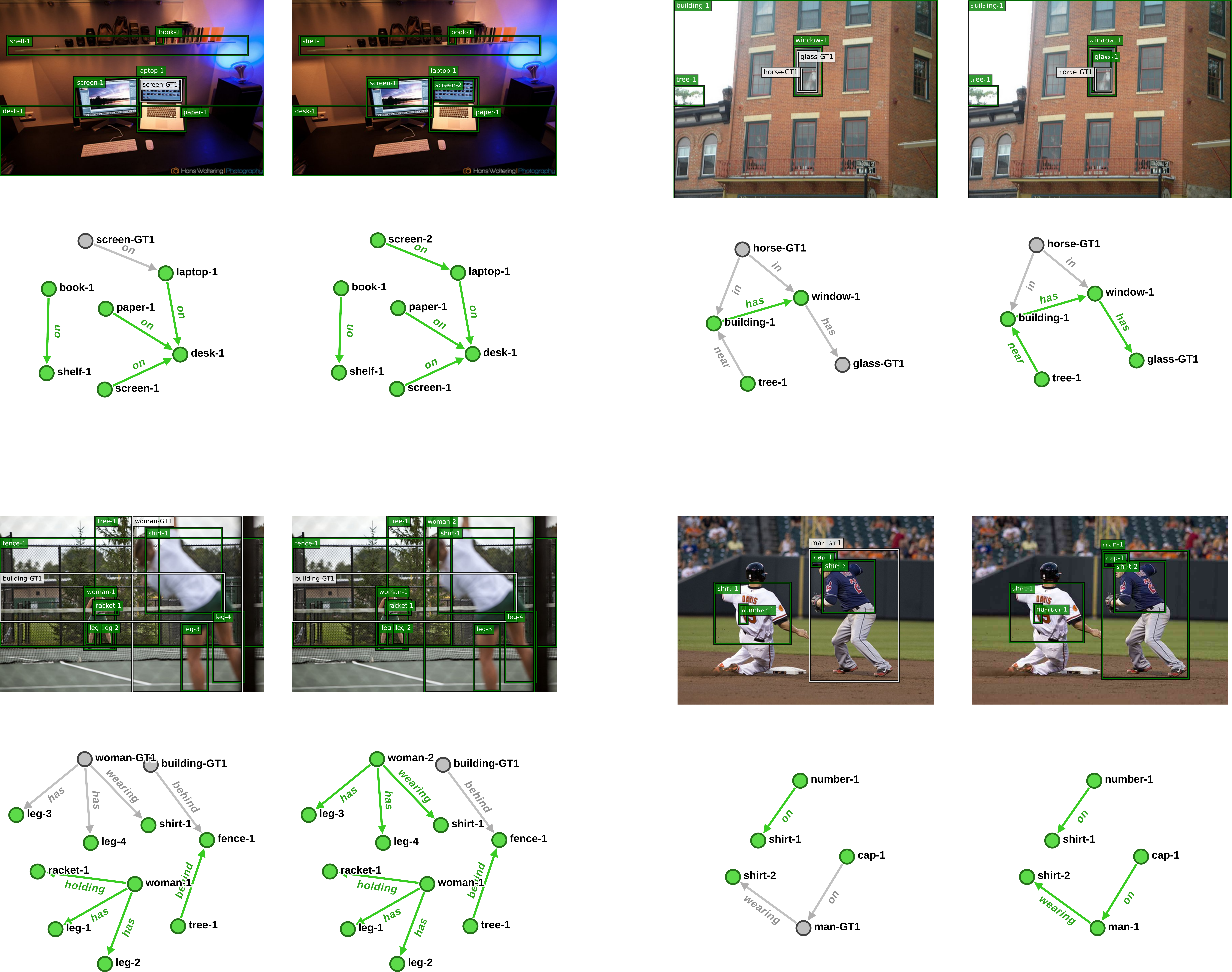}
			\end{center}
			\caption{Qualitative examples of improved scene graph classification results (R@50 - with graph constraints) through assimilations of our model. From left to right is after each assimilation. Green and gray colors indicate true positives and false negatives concluded by the model.}\label{fig_qual2}
		\end{figure*}

		In the main paper, we reported the per-predicate improvement of assimilation 1 to 8 in SGCls, under the R@100, and in the unconstrained setting. Here we present the additional results under R@50 and also in the constrained settings in Figure \ref{fig_perpred1}, \ref{fig_perpred2} and \ref{fig_perpred3}. The predicate statistics in all settings are the same and equal to the reported statistics in the main paper (Figure 7). Therefore, to avoid repeating the statistics under each plot and have a more visually appealing design, we embed these statistics as the brightness level of the bar colors. The darker shades indicate larger statistics and the lighter shades indicate smaller statistics. As we can see, most improvements occur in under-represented predicate classes.
	
		{\renewcommand{\arraystretch}{1.2}
			\begin{table*}
				\centering
				\small
				\scalebox{1.0}{
					\begin{tabular}{c|c|c|cc|cc}
						\hline
						\multirow{2}{*}{} &\centering \multirow{2}{*}{Method} &
						\centering \multirow{2}{*}{Main Model} &
						\multicolumn{2}{c|}{R@50} & \multicolumn{2}{c}{R@100}  \\
						& \multirow{2}{*} & & $1{\%}$ & $10{\%}$ & $1{\%}$ & $10{\%}$\\
						\hline
						\hline
						\multirow{3}{*}{\rotatebox[origin=c]{0}{SGCls}} 
						& \centering Supervised & \centering IC &  $\phantom{0}1.98\pm0.28$ & $16.24\pm1.13$ & $\phantom{0}2.40\pm0.30$& $18.15\pm1.23$  \\
						
						& \centering Self-Supervised & \centering IC &  $13.34\pm0.51$ & $30.45\pm0.95$ & $15.75\pm0.64$& $33.94\pm0.94$ \\
						& \centering \textbf{Self-Supervised} & \centering \textbf{IC + ICP} & $\mathbf{17.54\pm0.43}$ & $\mathbf{31.72\pm0.61}$ & $\mathbf{19.59\pm0.49}$& $\mathbf{35.33\pm0.65}$\\
						
						\hline
						
						\multirow{3}{*}{\rotatebox[origin=c]{0}{PredCls}} 
						& \centering Supervised & \centering IC &  $42.46\pm0.99$ & $60.21\pm0.93$ & $55.08\pm1.10$& $71.87\pm0.84$ \\
						
						& \centering Self-Supervised & \centering IC&  $51.77\pm0.59$ & $65.89\pm0.42$ & $63.64\pm0.75$& $76.85\pm0.38$ \\
						& \centering \textbf{Self-Supervised} & \centering \textbf{IC + ICP} &  $\mathbf{73.89\pm0.19}$ & $\mathbf{73.25\pm0.23}$ & $\mathbf{84.30\pm0.16}$& $\mathbf{83.69\pm0.20}$ \\
						
						\hline
				\end{tabular}}
				\caption{Comparison of R@50 and R@100 with no graph constraints, for SGCls and PredCls tasks on the provided VG splits.}
				\label{unconst_table_m}
			\end{table*}
		}
		{\renewcommand{\arraystretch}{1.2}
			\begin{table*}
				\centering
				\small
				\scalebox{1.0}{
					\begin{tabular}{c|c|c|cc|cc}
						\hline
						\multirow{2}{*}{} &\centering \multirow{2}{*}{Method}&
						\centering \multirow{2}{*}{Main Model} & \multicolumn{2}{c|}{mR@50} & \multicolumn{2}{c}{mR@100}  \\
						& \multirow{2}{*} & & $1{\%}$ & $10{\%}$ & $1{\%}$ & $10{\%}$\\
						\hline
						\hline
						\multirow{3}{*}{\rotatebox[origin=c]{0}{SGCls}} 
						& \centering Supervised &  \centering IC& $0.29\pm0.07$ & $4.65\pm0.62$ & $ 0.43\pm0.10$ & $\phantom{0}6.72\pm0.77$   \\
						
						& \centering Self-Supervised &  \centering IC& $2.54\pm0.42$ & $9.93\pm0.35$ & $3.71\pm0.49$ & $13.89\pm0.51$  \\
						& \centering \textbf{Self-Supervised} & 
						\centering \textbf{IC + ICP}&  $\mathbf{4.84\pm0.13}$ & $\mathbf{9.94\pm0.54}$ & $\mathbf{6.83\pm0.16}$ & $\mathbf{14.11\pm0.64}$ \\
						
						\hline
						
						\multirow{3}{*}{\rotatebox[origin=c]{0}{PredCls}} 
						& \centering Supervised & \centering IC&  $\phantom{0}5.89\pm0.26$ & $15.02\pm1.37$ & $\phantom{0}9.62\pm0.38$ & $23.22\pm1.78$  \\
						
						& \centering Self-Supervised & \centering IC&  $\phantom{0}9.24\pm1.31$ & $20.52\pm0.53$ & $14.28\pm1.63$ & $30.24\pm0.72$  \\
						& \centering \textbf{Self-Supervised} & 
						\centering \textbf{IC + ICP} &  $\mathbf{27.11\pm0.44}$ & $\mathbf{25.76\pm0.74}$ & $\mathbf{40.35\pm0.77}$ & $\mathbf{38.48\pm0.82}$  \\
						
						\hline
						
				\end{tabular}}
				\caption{Comparison of mR@50 and mR@100 with no graph constraints, for SGCls and PredCls tasks on the provided VG splits.}
				\label{unconst_table_mr}
			\end{table*}
		}
		{\renewcommand{\arraystretch}{1.2}
			\begin{table*}
				\centering
				\small
				\scalebox{1.0}{
					\begin{tabular}{c|c|c|cc|cc}
						\hline
						\multirow{2}{*}{} &\centering \multirow{2}{*}{Method}&
						\centering \multirow{2}{*}{Main Model} & \multicolumn{2}{c|}{R@50} & \multicolumn{2}{c}{R@100}  \\
						& \multirow{2}{*} & & $1{\%}$ & $10{\%}$ & $1{\%}$ & $10{\%}$\\
						\hline
						\hline
						\multirow{3}{*}{\rotatebox[origin=c]{0}{SGCls}} 
						& \centering Supervised & \centering IC&  $\phantom{0}1.65\pm0.26$ & $13.37\pm0.94$ & $\phantom{0}1.84\pm0.26$ & $13.90\pm0.97$   \\
						
						& \centering Self-Supervised & \centering IC& $11.19\pm0.41$ & $25.16\pm0.79$  & $12.12\pm0.47$ & $26.14\pm0.77$ \\
						& \centering \textbf{Self-Supervised} &
						\centering \textbf{IC + ICP} &  $\mathbf{14.72\pm0.38}$ & $\mathbf{26.33\pm0.45}$ & $\mathbf{15.36\pm0.38}$ & $\mathbf{27.37\pm0.47}$ \\
						
						\hline
						
						\multirow{3}{*}{\rotatebox[origin=c]{0}{PredCls}} 
						& \centering Supervised & \centering IC&  $34.92\pm0.81$ & $48.69\pm1.24$ & $40.61\pm0.84$  & $52.51\pm1.19$ \\
						
						& \centering Self-Supervised & \centering IC& $43.13\pm0.59$ & $54.40\pm0.39$  &  $48.10\pm0.54$ & $58.14\pm0.35$  \\
						& \centering \textbf{Self-Supervised} &
						\centering \textbf{IC + ICP}&  $\mathbf{62.02\pm0.10}$ & $\mathbf{61.71\pm0.19}$ & $\mathbf{65.68\pm0.12}$ & $\mathbf{65.42\pm0.19}$ \\
						
						\hline
						
				\end{tabular}}
				\caption{Comparison of R@50 and R@100 with graph constraints, for SGCls and PredCls tasks on the provided VG splits.}
				\label{table_r}
			\end{table*}
		}
		{\renewcommand{\arraystretch}{1.2}
			\begin{table*}
				\centering
				\small
				\scalebox{1.0}{
					\begin{tabular}{c|c|c|cc|cc}
						\hline
						\multirow{2}{*}{} &\centering \multirow{2}{*}{Method}&
						\centering \multirow{2}{*}{Main Model} & \multicolumn{2}{c|}{mR@50} & \multicolumn{2}{c}{mR@100}  \\
						& \multirow{2}{*}	& & $1{\%}$ & $10{\%}$ & $1{\%}$ & $10{\%}$\\
						\hline
						\hline
						\multirow{3}{*}{\rotatebox[origin=c]{0}{SGCls}} 
						& \centering Supervised & \centering IC&  $0.19\pm0.04$ & $2.21\pm0.34$ & $0.22\pm0.04$ & $2.43\pm0.36$  \\
						
						& \centering Self-Supervised & \centering IC&  $1.53\pm0.20$ & $\mathbf{5.31\pm0.39}$  &  $1.71\pm0.22$ & $\mathbf{5.80\pm0.40}$ \\
						& \centering \textbf{Self-Supervised} &
						\centering \textbf{IC + ICP}&  $\mathbf{2.45\pm0.05}$ & $5.17\pm0.23$    & $\mathbf{2.68\pm0.06}$ & $5.68\pm0.23$ \\
						
						\hline
						
						\multirow{3}{*}{\rotatebox[origin=c]{0}{PredCls}} 
						& \centering Supervised & \centering IC&  $\phantom{0}3.80\pm0.20$ & $\phantom{0}8.16\pm0.67$ & $\phantom{0}4.56\pm0.22$  & $\phantom{0}9.45\pm0.76$ \\
						
						& \centering Self-Supervised & \centering IC& $\phantom{0}5.64\pm0.56$ & $11.35\pm0.34$  &  $\phantom{0}6.61\pm0.62$ & $12.96\pm0.32$ \\
						& \centering \textbf{Self-Supervised} &
						\centering \textbf{IC + ICP}&  $\mathbf{14.80\pm0.31}$ & $\mathbf{14.59\pm0.60}$ & $\mathbf{17.06\pm0.39}$ & $\mathbf{16.92\pm0.58}$ \\
						
						\hline
						
				\end{tabular}}
				\caption{Comparison of mR@50 and mR@100 with graph constraints, for SGCls and PredCls tasks on the provided VG splits.}
				\label{table_m}
			\end{table*}
		}
		
		{\renewcommand{\arraystretch}{1.2}
			\begin{table*}
				\centering
				\small
				\scalebox{1.0}{
					\begin{tabular}{c|c|cc}
						\hline
						\centering \multirow{2}{*}{Method} &
						\centering \multirow{2}{*}{Main Model}& \multicolumn{2}{c}{Object Classification}  \\
						& \multirow{2}{*} & $1{\%}$ & $10{\%}$ \\
						\hline
						\hline
						\centering Supervised &\centering IC&  $14.38\pm0.57$ & $38.45\pm1.21$  \\
						
						\centering Self-Supervised & \centering IC& $40.75\pm0.48$ & $56.97\pm0.76$ \\
						\centering \textbf{Self-Supervised} &
						\centering \textbf{IC + ICP}&  $\mathbf{42.09\pm0.65}$& $\mathbf{58.60\pm0.56}$ \\

						\hline
				\end{tabular}}
				\caption{Comparison of object classification accuracy (Top-1) on the provided VG splits.}
				\label{table_classification}
			\end{table*}
		}


	
\end{document}